\documentclass[final,12pt]{colt2022} 

\usepackage{pifont}
\usepackage{graphicx}
\usepackage{rotating}
\usepackage{appendix}
\usepackage{algorithm, algorithmic}

\usepackage{amsmath,amsfonts,bm}

\def\eqref#1{equation~\ref{#1}}

\def\floor#1{\lfloor #1 \rfloor}
\def\1{\bm{1}}

\def\ra{{\textnormal{a}}}

\def\rs{{\textnormal{s}}}

\def\rva{{\mathbf{a}}}

\def\rvs{{\mathbf{s}}}

\def\va{{\bm{a}}}

\def\vs{{\bm{s}}}

\def\vpi{{\boldsymbol{\pi}}}

\def\mA{{\bm{A}}}

\def\mS{{\bm{S}}}

\DeclareMathAlphabet{\mathsfit}{\encodingdefault}{\sfdefault}{m}{sl}
\SetMathAlphabet{\mathsfit}{bold}{\encodingdefault}{\sfdefault}{bx}{n}

\newcommand{\E}{\mathbb{E}}

\DeclareMathOperator{\sign}{sign}

\usepackage{comment}
\usepackage{lipsum}
\usepackage{xcolor}
\newcommand{\x}{\mathbf{x}}
\newcommand{\w}{\mathbf{w}}

\newcommand{\tv}{\mathbf{v}}
\newcommand{\tu}{\mathbf{u}}

\newcommand{\tb}{\mathbf{b}}

\newcommand{\ta}{\mathbf{a}}

\newcommand{\tot}{\mathrm{tot}}
\newcommand{\dist}{\mathrm{dist}}
\newcommand{\Proj}{\mathrm{Proj}}
\newcommand{\ba}{\mathbf{a}}
\newcommand{\redc}[1]{\textcolor{red}{#1}}

\newtheorem{Theorem}{Theorem}[section]
\newtheorem{Lemma}{Lemma}[section]
\newtheorem{Proposition}{Proposition}[section]

\newtheorem{Assumption}{Assumption}[section]
\newtheorem{Definition}{Definition}[section]
\newtheorem{Remark}{Remark}[section]
\newenvironment{smalleralign}[1][\small]
 {\par\nopagebreak\leavevmode\vspace*{-\baselineskip}%
  \skip0=\abovedisplayskip
  #1%
  \def\maketag@@@##1{\hbox{\m@th\normalfont\normalsize##1}}%
  \abovedisplayskip=\skip0
  \align}
 {\endalign\ignorespacesafterend}
 
\title[Understanding Value Decomposition Algorithms in  Deep Cooperative MARL]{Understanding Value Decomposition Algorithms in \\ Deep Cooperative Multi-Agent Reinforcement Learning}
\usepackage{times}

\coltauthor{%
 \Name{Zehao Dou} \Email{zehao.dou@yale.edu}\\
 \addr Department of Statistics and Data Science, Yale University
 \AND
 \Name{Jakub Grudzien Kuba} \Email{jakub.grudzien@new.ox.ac.uk }\\
 \addr Department of Statistics, University of Oxford%
 \AND 
  \Name{Yaodong Yang} \Email{yaodong.yang@pku.edu.cn}\\
 \addr Institute for AI, Peking University \& BIGAI%
 \AND
}

\begin{document}

\maketitle
\allowdisplaybreaks[4]
\begin{abstract}%
Value function decomposition is becoming a popular rule of thumb for scaling up multi-agent reinforcement learning (MARL) in cooperative games. 
For such a decomposition rule to hold, the assumption of the \textit{individual-global max} (IGM) principle must be made; that is, the local maxima on the decomposed value function per every agent must amount to the global maximum on the joint value function. 
This principle, however, does not have to hold in general. As a result,  the applicability of value decomposition algorithms is concealed and  their corresponding convergence properties remain  unknown. In this paper, we make the first effort to answer these questions. 
Specifically, we introduce the set of cooperative games in which the value decomposition methods find their validity, which is referred as \textit{decomposable games}. In decomposable games, we  theoretically prove that applying the multi-agent fitted Q-Iteration algorithm (MA-FQI) will lead to an optimal Q-function. 
In non-decomposable games,  the estimated Q-function by MA-FQI can still converge to the optimum under the circumstance that the Q-function needs projecting into the decomposable function space at each iteration. 
In both settings, we consider value function representations by practical deep neural networks  and derive their corresponding convergence rates. 
To summarize, our results, for the first time, offer theoretical insights for MARL practitioners in terms of when value decomposition algorithms converge and why they perform well.  


\end{abstract}

\vspace{5pt}

\section{Introduction}

Q-learning is  one of the most classical approach of solving Markov decision processes in single-agent reinforcement learning (RL) \citep{sutton2018reinforcement}. At every iteration, a learning agent fits the state-action value critic, and then acts to maximize it. This method, combined with the expressive power of deep neural networks, enabled RL agents to learn to solve complex decision-making problems \citep{mnih2015human, silver2016mastering}. Although this hybrid approach serves as the template for designing deep RL methods, the difficulty of analyzing deep neural network models makes the understanding of it still lacking. However, recently, the first steps towards demystifying it were made by \cite{fan2020theoretical}, who derive the convergence rate of \textit{Deep Q-Network} (DQN) \citep{mnih2015human}. Their analysis uncovered that one of the keys behind the success of DQN is over-parameterization of the critic network, thus bridging the Q-learning framework and the deep-learning components of the algorithm.

In multi-agent reinforcement learning (MARL) \citep{yang2020overview},  applying effective Q-learning based method is no longer straightforward. If agents act independently, greedy policies with respect to their local state-action value functions do not necessarily maximize the global value function. This impedes agents from performing the policy improvement step of Q-learning. To tackle this issue, the framework of \textit{Centralized Training with Decentralized Execution (CTDE)} was introduced \citep{foerster2018counterfactual,pr2,gr2}. In the CTDE framework, the agents have access to the global critic during training, which enables the improvement of the joint policy. Afterwards, upon execution (once the training has ended), the learned joint critic is no longer accessible. Therefore, naively relying on the joint critic, not having learned adequate decentralized ones, the agents are put back at the starting point, let alone the large variance issue  \citep{kuba2021settling,kuba2021trust}.

A possible solution to the above issue is through enforcing the individual-global max (IGM) principle \citep{sunehag2017value, rashid2018qmix, yang2020multi} within the CTDE framework. IGM states that the global maximizer of the joint state-action value function is the concatenation of the maximizers of the agents' individual value components. The agents learn their local value functions with Q-learning by combining them monotonically to form the joint value function estimate. As we show in this paper, although this approach can work well in practice \citep{mahajan2019maven}, value function decompositions derived from the IGM principle do not hold in general. The lack of their generality increases the difficulty of their analysis and may have impeded us from demystifying the keys behind their empirical success, as well as the methods' limitations.

This work takes the first step towards understanding the state-of-the-art value-based algorithms. Its purpose is to describe the settings in which these algorithms can be employed, and settle their properties in these settings. With this aim, we first derive the set of cooperative games in which the value decomposition methods find their validity, which is referred as \textit{decomposable games}. Within the decomposable games, we then prove that applying the multi-agent fitted Q-Iteration algorithm (MA-FQI) can in fact lead to the optimal Q-function. 
This result offers theoretical insights for MARL practitioners in terms of when value decomposition algorithms converge and why they perform well.  
The second part of our contribution lies in the 
non-decomposable games,  wherein we show that the estimated Q-function by MA-FQI can still converge to the optimum, despite the fact that the estimated Q-function needs projecting into the decomposable function space at each iteration. 
In both decomposable and non-decomposable games, we consider value function representations by over-parameterized deep neural networks  and derive the corresponding convergence rates for MA-FQI.  Our work fills the research gap by providing theoretical insights, in terms of convergence guarantee, for the popular value decomposition algorithms in cooperative MARL. 



\section{Preliminaries \& Background}
In this section, we provide the background for MARL by introducing the fundamental definitions, and surveying the most important solution approaches. In Subsection \ref{sec-2.1}, we introduce the basic nomenclature for Markov Games, and in Subsection \ref{sec:value-based-algos}, we review value decomposition algorithms, such as VDN and QMIX. In Subsection \ref{sec-2.3}, we review the multi-agent fitted Q-iteration (MA-FQI) framework, which is the multi-agent version of the widely known FQI method. 
\subsection{Multi-Agent Markov Games}
\label{sec-2.1}
We start by defining the \textit{cooperative multi-agent Markov games} (MAMG) \citep{littman1994markov}. Formally, we consider the tabular episodic framework of the form $\mathcal{MG}(N, \boldsymbol{\mathcal{S}}, \boldsymbol{\mathcal{A}}, \mathbb{P}, R, \gamma, \pi_0)$, where 
$N$ is the number of agents (a.k.a players).
$\mathcal{S}^{(i)}$ is the state space of agent $i\in[N]$; without loss of generality, $\mathcal{S}=[0,1]^d$ for $d\in\mathbb{N}$. We write $\boldsymbol{\mathcal{S}}\triangleq \mathcal{S}^{(1)} \times \dots \times \mathcal{S}^{(N)}$ to denote the joint state space.
$\mathcal{A}^{(i)}$ is the action space of player $i$ and $\boldsymbol{\mathcal{A}}\triangleq \mathcal{A}^{(1)} \times \dots \times \mathcal{A}^{(N)}$ denotes the joint action space.
$\mathbb{P}$ is the transition probability function, so that $\mathbb{P}(\cdot|\vs,\va)$ gives the distribution over states if the joint action $\va= (a^1, \cdots,a^N)$ is taken at the (joint) state $\vs=(s^1, \dots, s^N)$.
$R:\boldsymbol{\mathcal{S}}\times \boldsymbol{\mathcal{A}}\rightarrow [-R_{\text{max}}, R_{\text{max}}]$ is the reward function. 
$\gamma \in [0, 1)$ is the discount factor.
$\pi_0$ is the initial state distribution.
In each episode of MAMG, an initial state $\rvs_0$ is drawn from the distribution $\pi_0$. Then, at every time step $t\in\mathbb{N}$, each player $i\in[N]$ observes the local state $\rs^i_t\in\mathcal{S}^{(i)}$, and takes an action $\ra^{i}_t\in\mathcal{A}^{(i)}$ according to its policy $\pi^{i}(\cdot^{i}|\rs^i_t)$, simultaneously with others. Equivalently, the agents take the joint action $\rva_t\in\boldsymbol{\mathcal{A}}$ at state $\rvs_t \in\boldsymbol{\mathcal{S}}$, according to their joint policy $\vpi(\cdot|\rvs_t) \triangleq \prod_{i\in[N]}\pi^i(\cdot^i|\rs^i_t)$. After that, the players receive the joint reward $R\big(\rvs_t, \rva_t\big)$ and transit to the next state $\rvs_{t+1}\sim\mathbb P\big(\cdot|\rvs_t, \rva_t\big)$.
We define the maximization objective of the collaborative agents, which is known as the joint return: 
\begin{align}
    \vspace{-0pt}
    \label{eq:joint-return}
    J(\boldsymbol{\pi}) \triangleq \E_{\rvs_0\sim\pi_0, \boldsymbol{a}_{0:\infty}\sim\boldsymbol{\pi}, \rvs_{1:\infty}\sim\mathbb{P}}\Big[ \sum\limits_{t=0}^{\infty}\gamma^t R(\rvs_t, \rva_t) \Big]. 
\end{align}
Crucially, as a proxy to the joint return, the agents guide their behavior with the joint state-action value function $Q^{\boldsymbol{\pi}}:\boldsymbol{\mathcal{S}}\times\boldsymbol{\mathcal{A}}\rightarrow [-\frac{R_{\max}}{1-\gamma}, \frac{R_{\max}}{1-\gamma}] \triangleq [-Q_{\text{max}}, Q_{\text{max}}]$, defined as
    \vspace{-0pt}
\begin{align}
    \label{eq:q-function}
    Q^{\boldsymbol{\pi}}(\vs, \boldsymbol{a}) \triangleq
    \E_{\rvs_{1:\infty}\sim\mathbb{P}, \rva_{1:\infty}\sim\boldsymbol{\pi}}\Big[ \sum\limits_{t=0}^{\infty}\gamma^t R(\rvs_t, \rva_t) \ | \ \rvs_0=\vs, \ \rva_0 = \boldsymbol{a}\Big],
\end{align}
on top of which one can define the state value function $V^{\pi}(\vs) \triangleq \E_{\rva\sim\vpi}\big[ Q^{\pi}(\vs, \rva) \big]$. In this paper, we are interested in the Q-learning type of approach to policy training. Ideally, at every iteration $k\in\mathbb{N}$, the agents would make the joint policy update $\boldsymbol{\pi}_{k+1}\big(\arg\max_{\va\in\boldsymbol{\mathcal{A}}}Q^{\boldsymbol{\pi}_k}(\vs, \va) | \vs\big) = 1$, 
\emph{i.e.}, act greedily with respect to $Q^{\boldsymbol{\pi}_k}$. 
Then, the sequence of state-action value functions $\{Q^{\boldsymbol{\pi}_k}\}_{k\in\mathbb{N}}$ would converge to the unique optimal joint state-action value function $Q^*$. The greedy joint policy, $\boldsymbol{\pi}^*\big(\arg\max_{\va\in\boldsymbol{\mathcal{A}}}Q^*(\vs, \boldsymbol{a}) | \vs\big) = 1$,
is the optimal joint policy, and maximizes the joint return $J(\boldsymbol{\pi}^*)$ \citep{sutton2018reinforcement}. Unfortunately, in MARL, the agents learn distributed (independent) policies. Therefore, even though the CTDE framework allows for implementation of the greedy joint policy during training, it does not scale to the execution phase. To circumvent this, a novel family of value decomposition methods has emerged, which we describe in the next subsection.

\subsection{Value Decomposition Algorithms}
\label{sec:value-based-algos}
We start by introducing the  pivotal IGM condition that value decomposition algorithms rely on; it enables global maximization in a decentralized manner. 
\begin{Definition}[IGM (Individual Global Max) Condition] 
\label{def:igm}
For a joint action-value function $Q_{\tot}:\boldsymbol{\mathcal{S}}\times\boldsymbol{\mathcal{A}}\rightarrow\mathbb{R}$, if there exist $N$ individual Q-functions $\{Q_i : \mathcal{S}^{(i)}\times\mathcal{A}^{(i)}\rightarrow\mathbb{R}\}$, such that: 
\begin{align}
    \arg\max_{\va\in\boldsymbol{\mathcal{A}}}Q_{\tot}(\vs, \ba)= \Big(
    \arg\max\limits_{a_1\in\mathcal{A}^{(1)}}Q_1(s^1,a^1), \ \dots, \ \arg\max\limits_{a^N\in\mathcal{A}^{(N)}}Q_N(s^N,a^N) \Big)
\end{align}
then $Q_{\text{tot}}$ satisfies the IGM condition with $\{Q_i\}_{i\in[N]}$ decomposition. If the IGM condition is met for all valid value functions $Q_{\text{tot}}$, then the MAMG is said to be decomposable.
\end{Definition}
As its name suggests, the condition states that individual optimal actions of the agents will constitute the optimal joint action. Without the IGM condition, we have to list all the $\sum_{i=1}^{N}|\mathcal{A}^{(i)}|$ possible joint actions in order to obtain the maximal $Q_{\tot}$ value. However, if the IGM condition holds, we only need to find the optimal action corresponding to the value function $Q_i$ for each $i\in[N]$, which only requires $\sum_{i=1}^{N}|\mathcal{A}^{(i)}|$ computational steps. Most crucially, this condition enables the agents to learn decentralized value functions which, once trained, can successfully be used in the execution phase. These potential benefits brought by ``decomposable" games invoke three  theoretical questions:
\textbf{1)} How to decide whether a game is decomposable?
\textbf{2)} How to find jointly optimal decentralized policies for decomposable games? \textbf{3)} How efficient the solutions to decomposable games are?

Although MARL researchers have not been indifferent about decomposability, they have only studied the problem via the second of the above questions. The first question would be skipped by an implicit assumption on the game's decomposability. Then, to tackle the second question, a solution to the game would be proposed, and its performance would be verified empirically \citep{sunehag2017value, rashid2018qmix, son2019qtran}. The last point remained ignored, leaving us without an idea of an explanation of the empirical efficacy of value decomposition algorithms. Nevertheless, the discovery of these methods is becoming a big step towards taming decomposable MARL problems. Below, we briefly introduce the first algorithm of this kind---VDN.

\paragraph{Value-Decomposition Network \citep[VDN]{sunehag2017value}} is a method which assumes that the global state-action value function satisfies the additive decomposition: for any $\vs\in \boldsymbol{\mathcal{S}}$, $\boldsymbol{a}\in \boldsymbol{\mathcal{A}}$,
    \vspace{-2pt}
\begin{align}
    \label{eq:vdn-assumption}
    Q_{\text{tot}}(\vs, \boldsymbol{a}) = \sum_{i=1}^{N}Q_i(s^i, a^i).
\end{align}
\vspace{-5pt}\\
The above structure implies that as, for any agent $i$, the value $Q_i(s^i, a^i)$ increases, so does $Q_{\text{tot}}(\vs, \boldsymbol{a})$. Hence, the IGM principle holds for any state-(joint)action pair, meaning that the game is decomposable. With this decomposition, VDN trains the decentralized critics by extending the Deep Q-Network (DQN) algorithm \citep{mnih2015human}. The greedy action selection with respect to $Q_{\text{tot}}$ step is performed by all agents $i$ acting greedily with respect to their local critics $Q_i$. Next, the critics are trained with TD-learning \citep{sutton2018reinforcement} with target networks, \emph{i.e.}, by minimizing
\begin{smalleralign}
    \label{eq:td-error}
    &\E_{\rvs, \rva, \rvs' \sim \mathcal{B}}\Bigg[ \Big( Q_\text{tot}(\rvs, \rva) - R(\rvs, \rva) - \gamma \max_{\boldsymbol{\rva'}}Q^{\text{tar}}_{\text{tot}}(\rvs', \boldsymbol{\rva'}) \Big)^2\Bigg] \nonumber\\
    &  \qquad \qquad  =
    \E_{\rvs, \rva, \rvs' \sim \mathcal{B}}\Bigg[ \Big( \sum\limits_{i=1}^{N}Q_i(\rs^i, \ra^i) - R(\rvs, \boldsymbol{a}) - \gamma \sum\limits_{i=1}^{N}\max_{\hat{\ra}^i}Q^{\text{tar}}_i(\rvs'^i, \rva'^i) \Big)^2\Bigg],
\end{smalleralign}
where $\mathcal{B}$ is the replay buffer. Intuitively, given empirical results from \cite{mnih2015human, sunehag2017value} and building upon the analysis of \cite{fan2020theoretical}, we should expect convergence guarantees of this algorithm, as long as the decomposition from Equation (\ref{eq:vdn-assumption}) is valid. In this paper, we affirm this intuition theoretically, and provide the key factors of the algorithm's efficacy.

One of the most popular extension of VDN is the QMIX algorithm \citep{rashid2018qmix}. The key novelty of the method is its general, IGM-compliant, value function decomposition,
\begin{align}
    \label{eq:qmix-assumption}
    Q_{\text{tot}}(\vs, \va) = \Phi(\vs)\big( Q_1(s^1, a^1), \dots, Q_N(s^N, a^N) \big). 
\end{align}
Here, for every $\vs\in \boldsymbol{\mathcal{S}}$, the function $\Phi(\vs):\mathbb{R}^{N}\rightarrow \mathbb{R}$ is the trainable \textit{mixing network}, whose weights are computed for every state by the network $\Phi(\cdot)$. Crucially, it satisfies the monotonicity assumption $\frac{\partial \Phi(\vs)(Q_1, \dots, Q_N)}{\partial Q_i} \geq 0$, which implies that the value of $Q_{\text{tot}}(\vs, \boldsymbol{a})$ increases monotonically with $Q_i(s^i, a^i)$. To guarantee this condition, the architecture of of the network $\Phi(\cdot)$ is constructed so that, for every state $s$, the weights of $\Phi(s)$ are non-negative. VDN is a special case of QMIX, with the mixing network taking form $\Phi^{\text{VDN}}(\vs)(Q_1, \dots, Q_N) = \sum_{i=1}^{N}Q_i$, for every state $\vs$. Monotonicity of QMIX, again, implies the IGM principle and decomposability of the game. Hence, the agents can learn their critics by acting greedily with respect to them, and repetitively minimizing the loss from Equation (\ref{eq:td-error}), substituting Equation (\ref{eq:qmix-assumption}) into $Q_{\text{tot}}$. As verified empirically, this method achieves substantially superior performance to that of VDN. However, there exist simple problems where QMIX fails utterly \citep{mahajan2019maven}. This warns us that the deployment of value decomposition algorithms, even those as powerful as QMIX, requires care and understanding. 

\subsection{Multi-Agent Fitted Q-Iteration (MA-FQI) Framework}
\label{sec-2.3}
Before we demystify the properties of the value decomposition algorithms, we specify the framework which generalizes all of them. Concretely, the core of these algorithms is the minimization of the (empirical) loss from Equation (\ref{eq:td-error}) within a function class $\mathcal{F}$. In practice, $\mathcal{F}$ is a family of neural networks with a specific architecture. The data $(\vs, \va)$ on which the minimization takes place is drawn from a large replay buffer $\mathcal{B}$. As argued by \cite{fan2020theoretical}, in the case of large state spaces and buffer sizes, independent draws of $(\vs, \va)$ from $\mathcal{B}$ constitute a marginal distribution $\sigma \in \mathcal{P}(\boldsymbol{\mathcal{S}} \times \boldsymbol{\mathcal{A}})$ which is fixed throughout training. These two steps of an empirical approximation and minimization of the squared TD-error is summarized by Algorithm \ref{algo:mafqi}---MA-FQI.
\begin{algorithm}[!ht]
\caption{Multi-Agent Fitted Q-Iteration Algorithm (MA-FQI)}
\label{algo:mafqi}
\textbf{Input:} MAMG $\mathcal{MG}(N, \boldsymbol{\mathcal{S}}, \boldsymbol{\mathcal{A}}, \mathbb{P}, R, \gamma, \pi_0)$, number of iterations $K$, function classes $\{\mathcal F_k\}_{k\in[K]}$, state-action sampling distribution $\sigma$, sample size $n$, initial Q-function estimate $\widetilde{Q}_0$.\newline
\vspace{-12pt}
\begin{algorithmic}[1]
\FOR {episode $k=0,1,2,\ldots,K-1$}
\STATE Sample i.i.d observations $\{(\vs_j, \va_j, R_j, \vs_j')\}_{j\in[n]}$ with $(\vs_j, \va_j)$ drawn from distribution $\sigma$. 
\STATE Compute targets $Y_j=R_j+\gamma \widetilde{Q}_k(\vs_j', a^1_*, \dots, a^N_*)$, where $\forall i\in[N]$, 
\vspace{-5pt}
\begin{align}
    &\quad \quad \quad \quad \quad \quad \quad \quad \quad  a^i_* = \arg\max_{a'^i\in\mathcal{A}^{(i)}}\widetilde{Q}^i_k(s_j'^i, a'^i) \quad \quad  \color{gray}\text{\textbackslash\textbackslash IGM condition}\color{black}\nonumber
\end{align}
\vspace{-10pt}
\STATE Update the joint action-value function:
$\widetilde{Q}_{k+1}\leftarrow \arg\min_{f\in\mathcal F_{k+1}}\frac{1}{n}\sum_{j=1}^{n}[Y_j-f(\vs_j, \va_j)]^2$.
\ENDFOR
\STATE Define the policy $\vpi_K$ as the product of the greedy policies $\{ \pi^i_K \}_{i\in[N]}$ with respect to $\{ \widetilde{Q}^i_K \}_{i\in[N]}$.
\end{algorithmic}
\hspace*{0.02in}{\bf Output:} An estimator $\widetilde{Q}_K$ of $Q^*$ and its greedy policy $\vpi_K$.
\end{algorithm}
\vspace{-5pt}

Compared with the Factorized Multi-Agent Fitted Q-Iteration (FMA-FQI) proposed by \cite{wang2021towards}, the state spaces are continuous thus infinite in our game setting, which is far beyond tabular case. Under the IGM condition, MA-FQI share certain similarities to its single-agent variant of FQI \citep{munos2008finite}: the step of computing targets through decentralized maximization gives the actual max-target, and the resulting distributed greedy policies result in a greedy joint policy. Hence, we can expect that the theoretical guarantees of FQI find their extension in MARL. Indeed, in the following sections, we show that the presence of multiple agents, does not prevent, yet slows down, the framework from convergence under the VDN model. 
\vspace{-5pt}

\section{Decomposable Games}
A preliminary step that we must take before we analyze the value decomposition algorithms is the analysis of frameworks that they are applicable to. Specifically, we characterize a class of MAMGs in which the additive value decomposition (i.e., Equation (\ref{eq:vdn-assumption})) holds.
\begin{Definition}[Decomposable Game] A multi-agent Markov Game $\mathcal{MG}(N, \boldsymbol{\mathcal{S}}, \boldsymbol{\mathcal{A}} , \mathbb{P}, R, \gamma, d_0)$ is a decomposable game if its reward function $R:\boldsymbol{\mathcal{S}}\times\boldsymbol{\mathcal{A}}\rightarrow \mathbb{R}$ can be decomposed as:
\[R(\vs, \va)=R_1(s^1, a^1)+R_2(s^2, a^2)+\ldots+R_N(s^N, a^N)\]
(here, $R_i:\mathcal{S}^{(i)}\times \mathcal{A}^{(i)}\rightarrow \mathbb{R}$ can be regarded as independent reward for the $i$-th agent) and the transition kernel $\mathbb{P}$ can be decomposed as:
\[\mathbb{P}(\vs'|\vs,\va) = F_1(\vs'|s^1, a^1) + F_2(\vs'|s^2, a^2) + \ldots + F_N(\vs'|s^N, a^N).\]
\end{Definition}

As we can see, a game is decomposable when both its reward function and its transition kernel can be distributed across individual agents and their local interactions with the game. The key property of a decomposable game is that the state-action value function $Q^{\vpi}$ can also be decomposed, regardless of the policy $\vpi$. This fact can be easily proved by expanding $Q^{\vpi}(\vs, \va)$ with the Bellman equation \citep{sutton2018reinforcement}:
\vspace{-10pt}
\begin{align}
Q^{\pi}(\vs, \va) &= R(\vs,\va) +\gamma\cdot \E_{\vs'\sim \mathbb{P}}\big[V^{\pi}(\vs')\big] = R(\vs,\va) + \gamma\int_{\boldsymbol{\mathcal{S}}}V^{\pi}(\vs') \mathbb{P}(\vs'|\vs,\va) d\vs'\nonumber\\
&=\sum_{i=1}^{N}R_i(s^i, a^i) + \gamma\int_{\boldsymbol{\mathcal{S}}} V^{\pi}(\vs')\cdot\left(\sum_{i=1}^{N}F_i(\vs'|s^i, a^i)\right)d\vs' \nonumber\\
&= \sum_{i=1}^{N}\left[R_i(s^i,a^i)+\gamma\int_{\boldsymbol{\mathcal{S}}} V^{\pi}(\vs')\cdot F_i(\vs'|s^i, a^i)d\vs'\right] \triangleq \sum_{i=1}^{N}Q^{\pi}_i(s^i, a^i).\nonumber
\end{align}    

Therefore, the decomposability of a game is a sufficient condition for the decomposability of the Q-value functions, which establishes the IGM principle in the game. In our studies, however, we pay most of our attention to the image of $Q$ under the Bellman operator $T$ \citep{sutton2018reinforcement} defined as $[TQ](\vs, \va) = R(\vs, \va) + \gamma\E_{\rvs'\sim\mathbb{P}}\big[ \max_{\va'}Q(\rvs', \va') \big]$,
because in Algorithm \ref{algo:mafqi} the critic $\widetilde{Q}_{k+1}$ is trained to match $T\widetilde{Q}_k$. Fortunately, in a decomposable game, $TQ$ is also decomposable. In fact, decomposable games are the \textbf{only} type of games in which this property holds, as given by the following proposition. 

\begin{Proposition}
For a MAMG $\mathcal{MG}( N,\boldsymbol{\mathcal{S}}, \boldsymbol{\mathcal{A}},\mathbb{P},R, \gamma, \pi_0)$, these two statements are equivalent:\\
(1) $\mathcal{MG}$ is a decomposable game.\\
(2) For any state-action value critic $Q$, and any discount factor $\gamma\in[0,1)$, $TQ$ is a decomposable function, i.e., there exist $G^{(\gamma)}_1, G^{(\gamma)}_2, \ldots, G^{(\gamma)}_N$ such that:
\begin{align}
\big[TQ\big](\vs,\va)=G^{(\gamma)}_1(s^1, a^1)+G^{(\gamma)}_2(s^2,a^2)+\ldots +G^{(\gamma)}_N(s^N, a^N)
\end{align}
\vspace{-10pt}
\label{proposition-1}
\end{Proposition}
See Appendix \ref{appendix:prop-decomposable} for proof. Hence, the algorithms which follow the framework of MA-FQI (Algorithm \ref{algo:mafqi}) implicitly make an assumption not simply about the decomposability of the Q-function, but also on the decomposability of the reward and transition functions. Although this setting might be rare in reality, it may be considered as its approximation through Taylor expansion up to the first order. The empirical success of VDN supports this point of view. Nevertheless, under this exact decomposable setting, we study the properties of VDN in the next section.

\section{Convergence Analysis in  Decomposable Games}
In this section, we study the convergence of VDN in the decomposable game. Precisely, we show that, in decomposable games, by considering the agents' joint action, VDN can be interpreted as DQN with a different (decomposed) function class. This similarity enables us to extend the analysis from single-agent deep RL \citep{munos2008finite} to MARL. 

We start by setting up the framework for function approximators. Firstly, for the purpose of convenience and clarity, we introduce the following assumptions.
\begin{Assumption}
All agents have identical state and action spaces, \emph{i.e.}, $\mathcal{S}^{(i)}=\mathcal{S}^{(j)}\triangleq\mathcal S$ and $\mathcal{A}^{(i)}=\mathcal{A}^{(j)}\triangleq\mathcal A$ hold for $\forall i, j\in[N]$.
\end{Assumption}
This assumption, although presented in practical applications, does not influence our analysis. It only enables us to simplify writing and notation, and allows us to replace quantities including summations with simple multiplication by the number of agents $N$. We proceed by defining the set of functions that are sums over maps of the decomposed input.
\begin{Definition}
Let $\mathcal{M}$ be a set of maps $m:\mathcal{S}\times\mathcal{A}\rightarrow\mathbb{R}$. Then, the $N$-composition set of $\mathcal{M}$ is defined as 
\vspace{-5pt}
\begin{align} 
    \mathcal{M}^{\oplus N} \triangleq \Big\{m^{(N)}:\boldsymbol{\mathcal{S}}\times\boldsymbol{\mathcal{A}}\rightarrow\mathbb{R} \ \big| \ m^{(N)}(\boldsymbol{s}, \boldsymbol{a}) = \sum\limits_{i=1}^{N}m^i(s^i, a^i), \ \text{and} \ \ m^i\in \mathcal{M}, \forall i\in[N]\Big\}.\nonumber
\end{align}
\end{Definition}
The role the above definition plays is that it captures the output of the joint critic of VDN into one function. It may be tempting to think that VDN simply adds $N$ decentralized and uncorrelated functions together, while the procedure of it is subtler. The algorithm, first, splits the state-action input $(\boldsymbol{s}, \boldsymbol{a})$ into $N$ parts, $\{(s^i, a^i)\}_{i\in[N]}$, then lets the parts pass through corresponding critics $\{Q^i\}_{i\in[N]}$, and computes their sum $Q_{\text{tot}}=\sum_{i=1}^{N}Q_i$ at the end. Thus, we can think of $Q_{\text{tot}}$ as of one joint critic, whose computation can be partially decentralized.

With this definition, and the intuition behind it, we continue our analysis. Crucially, as the joint critic $Q_{\text{tot}}$ is an element of an $N$-composition set, we must be able to study such sets. In particular, covering numbers of function classes play the key role in our considerations. One way to settle them is to take advantage of studies of neural networks, by relating the covering number of the $N$-composition set to that of its components, which we do in the following lemma.
\begin{Lemma}
\label{lemma:covering-n-sum}
Let $\mathcal{N}\left(\mathcal{M}, \delta\right)$ denote the cardinality of the minimal $\delta$ covering of set $\mathcal{M}$ of maps $m:\mathcal{S}\times \mathcal{A}\rightarrow\mathbb{R}$. Then we have
\begin{align}
    \mathcal{N}\left(\mathcal{M}^{\oplus N}, \delta\right)
    \leqslant \mathcal{N}\left(\mathcal{M}, \delta/N\right)^N.\nonumber
\end{align}
\end{Lemma}
For proof see Appendix \ref{appendix:proofs-N-compositions}.
Furthermore, we need a notion to describe the discrepancy between two function classes $\mathcal{M}_1$ and $\mathcal{M}_2$. In our analysis, most of the time we will need to study the worst-case scenario, of the mismatch between an approximator $m_1\in\mathcal{M}_1$ and the ground-truth $m_2\in\mathcal{M}_2$ be maximal possible. Therefore, we use the following notion of distance.
\begin{Definition}
 \label{def:sup-inf-distance}
    Let $\mathcal{M}_1$ and $\mathcal{M}_2$ be two classes of bounded functions with the domain $\mathcal{S}\times\mathcal{A}$ and image $\mathbb{R}$. Then, the distance between $\mathcal{M}_1$ and $\mathcal{M}_2$ is defined as
    \begin{align}
        \dist(\mathcal{M}_1, \mathcal{M}_2) \triangleq \sup_{m_1\in\mathcal{M}_1}\inf_{m_2\in\mathcal{M}_2}||m_1 - m_2||_{\infty}. \nonumber
    \end{align}
\end{Definition}
As before, having more control over the particular components $\{Q_i\}_{i\in[N]}$ of $Q_{\text{tot}}$, we are interested in relating the distance of two $N$-composition sets to the distance between their components. In the following lemma we obtain an elegant linear relation, derived in Appendix \ref{appendix:proofs-N-compositions}.
\begin{Lemma}
\label{lemma:bounding-distance} Let $\mathcal{M}_1$ and $\mathcal{M}_2$ be instances of function classes from Definition \ref{def:sup-inf-distance}. Then
\begin{align}
    \dist\big(\mathcal{M}_1^{\oplus N}, \mathcal{M}_2^{\oplus N} \big) \leqslant N\cdot\dist(\mathcal{M}_1, \mathcal{M}_2).\nonumber
\end{align}
\end{Lemma}

Knowing the relations between distributed functions and their $N$-composition, we possess tools that can unroll the properties of DQN to the multi-agent VDN algorithm. To give exact bounds, however, we must specify precisely the function approximators that the agents use. As it is often implemented by deep MARL practitioners, we let  every agent train a deep neural network, and the resulting joint critic is a summation over them.
\begin{Definition}[Deep Sparse ReLU Network]
    \label{def:sparse-nn}
    For any depth $L\in\mathbb{N}$, sparsity parameter $s\in\mathbb{N}$, and sequence of widths $\{d_j\}_{j=0}^{L+1}\subseteq \mathbb{N}$, and $U>0$, the function class $\mathcal{F}\left(L, \{d_j\}_{j=0}^{L+1}, s, U\right)$ is the set of maps $f:\mathbb{R}^{d_0}\rightarrow\mathbb{R}^{d_{L+1}}$, defined as
    \begin{align}
        f(x) = W_{L+1}\sigma\big( W_L\sigma(\dots(W_2\sigma(W_1x + v_1) + v_2)\dots v_{L-1}) + v_L\big), \nonumber
    \end{align}
    where for $j=0, \dots, L$, $W_{j+1} \in\mathbb{R}^{d_{j+1}\times d_j}$ are weight matrices, $v_j$ are bias vectors, and $\sigma(x) = \max(0, x)$ is the ReLU activation function. Furthermore, for this class we require that the weights of the network are not too large, i.e., $||(W_l, v_l)||_{\max}\leqslant 1$, $\forall l\in[L+1]$, not too many of the weights are non-zero, i.e, $\sum_{l=1}^{L+1}||(W_l, v_l)||_{\max} \leqslant s$, and that $\max_{j\in d_{L+1}} ||f_j||_{\infty} \leq U$. 
\end{Definition}
In our analysis, the efficacy of a network is related to its smoothness properties. To study them, we introduce the notion of H{\" o}lder smoothness---a tool considered in deep learning and reinforcement learning literature \citep{chen2019information, fan2020theoretical}.
\begin{Definition}[H{\" o}lder Smooth Function]
Let $d\in\mathbb{N}$, and $\mathcal{D}\subset\mathbb{R}^d$ be a compact set, and let $\beta, B>0$. The H{\" o}lder smooth functions on $\mathcal{D}$ are elements of the set
\begin{smalleralign}
    \mathcal{C}_d(\mathcal{D}, \beta, B) = \Bigg\{ f:\mathcal{D}\rightarrow\mathbb{R} \ : \ \sum_{|\boldsymbol{\alpha}|<\beta}||\partial^{\boldsymbol{\alpha}}f||_{\infty} + \sum_{||\boldsymbol{\alpha}||_1\leqslant \floor{\beta}}\sum_{x\neq y \in\mathcal{D}}\frac{ |\partial^{\boldsymbol{\alpha}}f(x) - \partial^{\boldsymbol{\alpha}}f(y)|}{||x-y||_{\infty}^{\beta - \floor{\beta}}}\leqslant B \Bigg\}, \nonumber
\end{smalleralign}
where $\floor{\beta}$ is the floor of $\beta$, $\boldsymbol{\alpha} = (\alpha_1, \dots, \alpha_d)\in\mathbb{N}^d$, and $\partial^{\boldsymbol{\alpha}} = \partial^{\alpha_1}\dots\partial^{\alpha_d}$.
\end{Definition}
Furthermore, to study fine compositions of real mappings, we must define a class of compositions of H{\"o}lder smooth functions.
\begin{Definition}[Composition of H{\"o}lder smooth functions]
\label{def:holder-smooth}
Let $q\in\mathbb{N}$ and $\{p_j\}_{j\in[q]}$ be integers, and let $\{[a_j, b_j]\}_{j\in[q]}$ non-empty real intervals. For any $j\in[q]$, consider a vector-valued function $g_j:[a_j, b_j]^{p_j}\rightarrow[a_{j+1}, b_{j+1}]^{p_{j+1}}$, such that each of its components $g_{j, k}$ ($k\in[p_{j+1}]$) is H{\"o}lder smooth, and depends on $t_j\leqslant p_j$ components of its input. We set $p_{q+1} =1$, and define the class of compositions of H{\"o}lder smooth functions $\mathcal{G}(\{p_j, t_j, a_j, b_j\}_{j\in[q]})$ as functions $f$, that can be written in a form $f= g_q \circ g_{q-1} \circ \cdots \circ g_1$,
where $g_1, \dots, g_q$ follow the rules listed above.
\end{Definition}
\label{definition:function-classes}
In our study, it is important  the type of neural network stays close to the above class, wherein their training targets happen to find themselves. Next, we lay out the characterization of the networks used by the agents, and that of compositions of H{\" o}lder smooth functions that the networks track.
\begin{Definition}[Function Classes]
Let $d_0 = d$, and $d_{L+1}= 1$. Then, any agent has access to the function class of neural network (Definition \ref{def:sparse-nn}) critics
\begin{align}
    \mathcal{F}_{\text{net}} \triangleq \Big\{ f:\mathcal{S}\times\mathcal{A}\rightarrow\mathbb{R} \ : \ f(\cdot, a)\in\mathcal{F}\left(L, \{d_j\}_{j=0}^{L+1}, s, Q_{\max}/N\right), \forall a\in\mathcal{A}\Big\}.\nonumber
\end{align}
Correspondingly, employing compositions of H{\"o}lder smooth functions (Definition \ref{def:holder-smooth}), we define the class
\begin{align}
    \mathcal{G}_{\text{H}} \triangleq \Big\{ g:\mathcal{S}\times\mathcal{A}\rightarrow\mathbb{R} \ : \ g(\cdot, a)\in\mathcal{G}\left( \{p_j, t_j, \beta_j, B_j\}_{j\in[q]}\right), \forall a\in\mathcal{A}\Big\},\nonumber
\end{align}
and refer to it as \textup{the H{\"o}lder class} for brevity. It follows that the joint critic $Q_{\text{tot}}$ belongs to the class $\mathcal{F}_{\text{net}}^{\oplus N}$, which tracks the corresponding H{\"o}lder class $\mathcal{G}_{\text{H}}^{\oplus N}$.
\end{Definition}

In the following, we make a standard assumption on approximate closure of $\mathcal{F}_{\text{net}}^{\oplus N}$ under the Bellman operator $T$, where the vicinity of $\mathcal{F}_{\text{net}}^{\oplus N}$ is considered to be $\mathcal{G}_{\text{H}}^{\oplus N}$ \citep{chen2019information, fan2020theoretical}. Note that, if the joint critic was able to learn the optimal value $Q^*$, then by the Bellman optimality equation $TQ^* = Q^*$. Hence, we would have $Q^*\in\mathcal{F}_{\text{net}}^{\oplus N}$ and $TQ^*\in\mathcal{F}_{\text{net}}^{\oplus N}$, which suggests the approximate closure.

\begin{Assumption}
For any $f\in\mathcal{F}_{\text{net}}^{\oplus N}$, we have $Tf\in\mathcal{G}^{\oplus N}_{\text{H}}$, where $T$ is the Bellman operator.
\end{Assumption}
Lastly, we make an assumption about the concentration coefficients \citep{munos2008finite}, which provide some notion of distance between two probability distributions on $\boldsymbol{\mathcal{S}}\times\boldsymbol{\mathcal{A}}$ in a MAMG.

\begin{Assumption}[Concentration Coefficients]
\label{assumption:concentration}
Let $\mathcal{P}(\boldsymbol{\mathcal{S}}\times\boldsymbol{\mathcal{A}})$ be the set of probability measures that are absolutely continuous with respect to the Lebesgue measure on $\boldsymbol{\mathcal{S}}\times\boldsymbol{\mathcal{A}}$.
Let $\nu_1, \nu_2\in\mathcal{P}(\boldsymbol{\mathcal{S}}\times\boldsymbol{\mathcal{A}})$, and the initial state-(joint)action pair has distribution $(\vs_0, \va_0)\sim\nu_1$. Let $\{\vpi_t\}_{t=1}^{\infty}$ be a sequence of joint policies so that, for $t\geq 1$, $\rva_t\sim\pi_t(\cdot|\rvs_t)$, and $\mathbb{P}^{\pi_t}\dots\mathbb{P}^{\pi_1}\nu_1$ is the marginal distribution of $(\rvs_t, \rva_t)$. We define the concentration coefficient at time $t$ as
\begin{smalleralign}
    \kappa_t(\nu_1, \nu_2) \triangleq \sup_{\pi_1, \dots, \pi_t}\left[
    \E_{\rvs, \rva\sim\nu_2}\left( \left|
    \frac{\text{d}\left(\mathbb{P}^{\pi_t}\dots \mathbb{P}^{\pi_1}\nu_1
    \right) }{\text{d}\nu_2} (\rvs, \rva)\right|^2 \right) \right]
    \nonumber
\end{smalleralign}
We assume that for $\nu_2=\sigma$, the sampling distribution of Algorithm \ref{algo:mafqi}, for any $\nu_1 = \mu\in\mathcal{P}(\boldsymbol{\mathcal{S}}\times\boldsymbol{\mathcal{A}})$, there exists a finite constant $\phi_{\mu, \sigma}$ such that $\phi_{\mu, \sigma} = (1-\gamma)^2 \sum\limits_{t=1}^{\infty}t\gamma^{t-1}\kappa_t(\mu, \sigma)$.
\end{Assumption}
With the definitions and assumptions set up, we finally reaching to discovering the theoretical properties of Algorithm \ref{algo:mafqi}.

\subsection{Theoretical Properties of Algorithm \ref{algo:mafqi}}
The following theorem  describes how the error in MA-FQI (Algorithm \ref{algo:mafqi}) propagates, and holds regardless of the function class used for critics. Extending the error propagation theorem for single-agent FQI to cooperative decomposable MAMGs, we have the following error bound.
\begin{Theorem}[Error Propagation]
\label{theorem:error-prop}
Let $K\in\mathbb{N}$, and $\{\widetilde{Q}_k\}_{k\in[K]}$ be iterates of Algorithm \ref{algo:mafqi} in a decomposable MAMG. Let $\pi_K$ be the greedy joint policy with respect to $\widetilde{Q}_K$, and $Q^{\pi_K}$ be the actual state-action value function of $\pi_K$. Recall that $Q^*$ denotes the optimal state-action value function. Under Assumption \ref{assumption:concentration}, 
\begin{align}
    \big\|Q^* - Q^{\pi_K}\big\|_{1, \mu} \leqslant \frac{ 2\phi_{\mu, \sigma}\gamma }{(1-\gamma)^2}\cdot \epsilon_{\max} + \frac{4\gamma^{K+1}}{(1-\gamma)^2}R_{\max},\ 
    \text{where} \ \epsilon_{\max} = \max_{k\in[K]}  \big\| T\widetilde{Q}_{k-1} - \widetilde{Q}_k\big\|_{\sigma}. \nonumber
\end{align}
\end{Theorem}

This theorem decomposes the final error into the \textsl{approximation error}, $2\phi_{\mu, \sigma}\gamma\epsilon_{\max}/(1-\gamma)^2$, and the \textsl{algorithmic error}, $4\gamma^{K+1}R_{\max}/(1-\gamma)^2$. The latter term does not depend on the function approximators used, and vanishes fast as the number of iterations $K$ increases. Therefore, the problem is the former term---the error arising from the function approximator, and in particular, its approximate closure under the Bellman operator $T$. Hereafter, we focus our analysis on it, which we begin with the following theorem, proved by \cite{fan2020theoretical}.
\begin{Theorem}[One-step Approximation Error]
\label{theorem:one-step-approx}
Let $\sigma$ be a probability distribution on $\boldsymbol{\mathcal{S}}\times\boldsymbol{\mathcal{A}}$, and let $\{(\vs_i, \va_i)\}_{i\in[n]}$ be a sample drawn from $\sigma$. Let $R_i$ and $\vs'_i$ be the reward and the next state corresponding to $(\vs_i, \va_i)$. Let $Q^{\text{tar}}\in\mathcal{F}^{\oplus N}_{\text{net}}$. For every $i\in[n]$, we define the training target $Y_i = R_i + \gamma \max_{\va\in\boldsymbol{\mathcal{A}}}Q^{\text{tar}}(\vs'_i, \va)$. Let
\begin{align}
    \hat{Q} = \arg\min_{f\in\mathcal{F}^{\oplus N}_{\text{net}}}\frac{1}{n}\sum\limits_{i=1}^{n}\big[ f(\vs_i, \va_i) - Y_i \big]^2,\nonumber
\end{align}
and for any $\delta>0$ and function class $\mathcal{F}$, let $\mathcal{N}(\delta, \mathcal{F}, ||\cdot||_{\infty})$ denote the cardinality of the minimal $\delta$-covering of $\mathcal{F}$, with respect to $l_{\infty}$-norm. Then, for some absolute constant $C>0$,
\begin{align}
    \big\|\hat{Q} - TQ^{\mathrm{tar}}\big\|^2_{\sigma} \leqslant 4\mathrm{dist}(\mathcal{F}^{\oplus N}_{\text{net}}, \mathcal{G}^{\oplus N}_{\text{H}})^2 + C\cdot(Q^2_{\max}/n)\cdot\log\mathcal{N}(\mathcal{F}^{\oplus N}_{\text{net}}, \delta, ||\cdot||_{\infty}).\nonumber
\end{align}
\end{Theorem}
The theorem decomposes the approximation error into quantities that are properties of the function approximator class and the (target) H{\"o}lder class. The first term, involving $\text{dist}(\mathcal{F}^{\oplus N}_{\text{net}}, \mathcal{G}^{\oplus N}_{\text{H}})$, can be thought of as a metric of mismatch between the class $\mathcal{F}^{\oplus N}_{\text{net}}$ and the class of targets $\mathcal{G}^{\oplus N}_{\text{H}}$. The better neural networks approximate the H{\"o}lder functions, the smaller this metric is. The second term, involving $\mathcal{N}(\mathcal{F}^{\oplus N}_{\text{net}}, \delta, ||\cdot||_{\infty})$, can be thought of as a measure of sparsity of the class. The less expressible the networks are, the bigger its $\delta$-covering. By considering sparse ReLU networks, and their $N$-composition that the agents use during learning, we provide the main theorem of this section, which reveals the convergence property of VDN in decomposable games.

\begin{Theorem}[Main Theorem 1: Decomposable Setting]
\label{theorem:deep-vdn-convergence}
Let $\mathcal{F}_{\text{net}}$ and $\mathcal{G}_{\text{H}}$ be defined as in Definition \ref{definition:function-classes}, based on the class of neural networks $\mathcal{F}_1=\ldots=\mathcal{F}_K=\mathcal{F}\left(L^*, \{d_j\}_{j=0}^{L^*+1}, s^*, Q_{\max}/N\right)$, and the class of H{\"o}lder smooth functions $\mathcal{G}_{\text{H}}\left( \{p_j, t_j, \beta_j, B_j\}_{j\in[q]}\right)$. For any $j\in[q-1]$, we define $\beta^*_j = \beta_j \prod_{l=j+1}^{q}\min(\beta_l, 1)$, and $\beta^*_q = 1$. In addition, let let $\alpha^* = \max_{j\in[q]}\frac{t_j}{2\beta^*_j + t_j} < 1$. We assume that the sample size is large, relative to the parameters of $\mathcal{G}_{\text{H}}$, so that there exists a constant $\xi>0$, such that
\vspace{-5pt}
\begin{align}
    \label{eq:hyperparams1}
    \max\Big\{ \sum\limits_{j=1}^{q}(t_j + \beta_j + 1)^{3+t_j}, 
    \sum_{j\in[q]}\log(t_j + \beta_j),
    \max_{j\in[q]}p_j\Big\} = \mathcal{O}\left( (\log n)^{\xi} \right).
\end{align}
Moreover, we assume that the hyper-parameters of the neural networks satisfy
\vspace{-5pt}
\begin{align}
    \label{eq:hyperparams2}
    L^* = \mathcal{O}\left( (\log n)^{\xi^*} \right), \ d\leqslant \min_{j\in[L^*]}d^*_j \leqslant \max_{j\in[L^*]}d^*_j = \mathcal{O}(n^{\xi^*}), \ \text{and} \ s^* = \Theta\left( n^{\alpha^*}(\log n)^{\xi^*} \right),
\end{align}
for some absolute constant $\xi^*>1+2\xi$. Let $\pi_K$ be the output joint policy of Algorithm \ref{algo:mafqi}, and $Q^{\pi_K}$ be its (true) joint state-action value function. Then, for some absolute constant $C>0$, 
\begin{align}
    &||Q^* - Q^{\pi_K}||_{1, \mu} \nonumber\\
    &\leqslant \frac{ C \phi_{\mu, \sigma} \gamma}{(1-\gamma)^2}\left( N\cdot n^{-(1- \alpha^*)/2} + \sqrt{|\mathcal{A}|\cdot N\cdot \log N}\cdot n^{-(1-\alpha^*)/2}(\log n)^{(1+2\alpha^*)/2} \right) + \frac{4\gamma^{K+1}}{(1-\gamma)^2}\cdot R_{\max}.\nonumber
\end{align}
\end{Theorem}
For proof see Appendix \ref{appendix:vdn-convergence}.
\vspace{-10pt}
\section{Convergence Analysis in Non-decomposable Games}
\vspace{-5pt}
\label{sec-non}
In this section, we extend the decomposable games to the general non-decomposable games, which is a more challenging setting. For simplicity, here instead of using multi-layer networks for training, we consider the 2-layer ReLU networks. Our proof can be easily applied to more  complicated function classes (such as multi-layer networks). Under this setting, we make the function class used in the $k$-th iteration to be $\mathcal F_k = \mathcal F(B_k, M)^{\oplus N}$, the set of decomposable 2-layer ReLU networks with weight $M$ and their path norm bounded by $B_k$, to be rigorous:
\vspace{-5pt}
{\small
\[\mathcal F(B,M)=\left\{f:\mathcal S\times\mathcal A\rightarrow\mathbb{R}\bigg|~f(s,a)=\sum_{i=1}^{M}\alpha_i^a\cdot\left(\langle \beta_i^a, s\rangle+\gamma_i^a\right),\max_{a\in\mathcal A}\sum_{i=1}^M |\alpha_i^a|\cdot\left(\|\beta_i^a\|_1+|\gamma_i^a|\right)
\leqslant B\right\},\vspace{-5pt}\]}
and $\mathcal F(B,M)^{\oplus N}\in\{F:\boldsymbol{\mathcal S}\times \boldsymbol{\mathcal A}\rightarrow\mathbb{R}\}$ is its $N$-composition. 
In the following parts, we are going to show that even for non-decomposable game where $T\widetilde{Q}_k$ may not be close to any decomposable functions for a decomposable $\widetilde{Q}_k$, the MA-FQI Algorithm will still be able to converge to the optimal value function $Q^*$ as long as $Q^*$ itself is a decomposable function, which is in fact a counterfactual result since we need to project our estimator onto the decomposable function class in each iteration, which may cause divergence by our intuition. In the following paragraphs, we are going to show that $\widetilde{Q}_k$ will provably converge to $Q^*$ when following Algorithm \ref{algo:mafqi}. First, we are going to bridge the gap between the value function of the greedy policy $\pi_k$, denoted by $Q^{\pi_k}$ and the estimated Q-value $\widetilde{Q}_k$ by the following lemma:
\vspace{-5pt}
\begin{Lemma}
\label{lemma-non-1}
\vspace{-3pt}
{\small\[\|Q^*-Q^{\pi_k}\|_{\infty}\leqslant \frac{2\gamma}{1-\gamma} \|Q^*-\widetilde{Q}_k\|_{\infty}.\]}
\end{Lemma}
\vspace{-3pt}
Therefore, in order to control the error $\|Q^*-Q^{\pi_k}\|_{\infty}$, we only need to upper bound the estimation error of Q-function, which is $\|Q^*-\widetilde{Q}_k\|_{\infty}$. Since $\widetilde{Q}_k$ is generated in an iterative manner and $\widetilde{Q}_{k+1}=\mathrm{Proj}\left(T\widetilde{Q}_k, \mathcal F(B_k)^{\oplus N}, \|\cdot\|_{\sigma}\right)$, we can upper bound the last iteration error $\|Q^*-\widetilde{Q}_K\|_{\infty}$ in a cumulative way.
\vspace{-5pt}
\begin{Lemma}
\label{lemma-non-2}
\vspace{-3pt}
{\small\[\big\|Q^*-\widetilde{Q}_K\big\|_{\infty}\leqslant \frac{\varepsilon_{\max}}{1-\eta}+\frac{4\gamma^K}{(1-\gamma)^2}R_{\max}.\vspace{-3pt}\]}
Here, $\eta=(N+1)\gamma$ and $\varepsilon_{\max}=\max_{k\in [K]}\left\|\widetilde{Q}_{k+1}-\Proj\left(T\widetilde{Q}_k, \mathcal C^{\oplus N}, \|\cdot\|_{\sigma}\right)\right\|_{\infty}$.
\end{Lemma}
From the two lemmas above, we know that in order to upper bound the gap $\|Q^*-Q^{\pi_K}\|_{\infty}$, we only need to upper bound the $\varepsilon_{\max}$. Next, we are going to prove that, with high probability over the sampling of $(s,a)\sim \sigma$ in each iteration, the discrepancy $\varepsilon_{\max}$ can be well upper bounded by using the approximation properties as well as the generalization properties of 2-layer ReLU networks (which are introduced in detail in Appendix \ref{appendix-e}). 
\begin{Lemma}
\label{lemma-non-3}
With probability at least $1-\delta$ over the sampling of  $(s,a)\sim \sigma$ in all the $K$ iterations, when the discount ratio $\gamma\ll\frac{1}{N^2}$, we can let $B_k = \frac{8Nc_2 R_{\max}}{1-4N^2\gamma}:=B$ for $\forall k\in[K]$, such that:
\vspace{-10pt}
{\small
\begin{equation}
\label{eqn-lemma-non-3-*}
\begin{aligned}
&\varepsilon_{\max}:= \max_{k\in[K]}\left\|\widetilde{Q}_{k+1}-\Proj\left(T\widetilde{Q}_k, \mathcal C^{\oplus N}, \|\cdot\|_{\sigma}\right)\right\|_{\infty} \leqslant c_1Bd\cdot\Bigg[\frac{B^2}{M}+8|\mA|\cdot\frac{\log(8|\mA|/\delta)}{n}\cdot Q_{\max}^2\notag\\
&~~+ \sqrt{|\mA|}\cdot\left(16Q_{\max}(2B+2)\sqrt{\frac{2\log(2d)}{n}}+8Q_{\max}^2\sqrt{\frac{8|\mA|\log(2c(B+1)^2/\delta)}{n}}\right)\Bigg]^{\frac{1}{d+2}},
\end{aligned}
\end{equation}
}
\vspace{-2pt}
where $c, c_1, c_2$ are constants. 
\end{Lemma}
Finally, after combining all the three lemmas above, we conclude that:
\begin{Theorem} [Main Theorem 2: Non-decomposable Setting]
\label{theorem:non-decomposable}
Assume the optimal Q-function $Q^*\in\mathcal C^{\oplus N}$ is a decomposable function. Let the function classes $\mathcal F_1 = \ldots = \mathcal F_K = \mathcal F(B,M)^{\oplus N}$. When the discount ratio $\gamma \ll\frac{1}{N^2}$, we can choose the path norm bound $B=\frac{8Nc_2 R_{\max}}{1-4N^2\gamma}$. Then, by running MA-FQI (Algorithm \ref{algo:mafqi}), for some constant $c, c_1, c_2 > 0$, we have:
\vspace{-8pt}
{\small
\begin{equation}
\label{eqn-lemma-non-3-final}
\begin{aligned}
&\|Q^*-\widetilde{Q}^{\pi_K}\|_{\infty}\leqslant \frac{8\gamma^{K+1}}{(1-\gamma)^3}R_{\max} +  \frac{c_1Bd\gamma}{(1-(N+1)\gamma)(1-\gamma)}\cdot\Bigg[\frac{B^2}{M}+8|\mA|\cdot\frac{\log(8|\mA|/\delta)}{n}\cdot Q_{\max}^2\notag\\
&~~+ \sqrt{|\mA|}\cdot\left(16Q_{\max}(2B+2)\sqrt{\frac{2\log(2d)}{n}}+8Q_{\max}^2\sqrt{\frac{8|\mA|\log(2c(B+1)^2/\delta)}{n}}\right)\Bigg]^{\frac{1}{d+2}}.
\end{aligned}
\end{equation}
}
\end{Theorem}
\vspace{-5pt}
For proofs see Appendix \ref{appendix-d}. As we can see, the first term $\frac{8\gamma^{K+1}}{(1-\gamma)^3}R_{\max}$ exponentially shrinks to 0 since $\gamma < 1$. For the second term, after treating all the instance-based parameters (such as $B,d,\gamma,N, Q_{\max}$) as constants, has order $\mathcal O\left(\frac1M\right)+\mathcal O\left(\frac{1}{\sqrt{n}}\right)$. Here, $\mathcal O\left(\frac1M\right)$ and $\mathcal O\left(\frac{1}{\sqrt{n}}\right)$ come from the approximation error and generalization error of 2-layer ReLU networks respectively. For sufficiently large width $M$ (which stands for the over-parameterization) and large sample size $n$ (which stands for the small gap between the empirical mean and population mean in the sampling process of each iteration), the $l_{\infty}$ error between $Q^*$ and $Q^{\pi_K}$ converges to 0. Although the sample complexity $\mathcal O(1/\varepsilon^{2d+4})$ suffers the curse of dimension, the convergence itself is a huge step for understanding the MA-FQI algorithm in cooperative multi-agent reinforcement learning. 
\vspace{-4pt}
\section{Conclusion}
\vspace{-4pt}
Although value decomposition methods for cooperative  MARL has great promise for addressing coordination problems in a variety of applications \citep{yang2017study,zhou2020smarts,zhou2021malib}, theoretical understandings for these  approaches are still limited. 
This paper makes the initial effort to bridge this gap by considering a general framework for theoretical studies. Central to our findings is the decomposable games where value decomposition methods can be applied safely. Specifically, we show that the multi-agent fitted Q-Iteration algorithm (MA-FQI), parameterized by multi-layer deep ReLU networks, can lead to the optimal Q-function. 
Moreover, for non-decomposable games,  the estimated Q-function parameterized by wide 2-layer ReLU networks, can still converge to the optimum by using MA-FQI, despite the fact that the Q-function needs projecting into the decomposable function space at each iteration. 
In our future works, we are going to extend the 2-layer ReLU networks to a much broader function class, and see whether we can reduce the sample complexity and avoid the curse of dimension. Also, mean-field game setting will be taken into consideration and we will see whether the convergence guarantee can still be provided in the sense of distribution. 

\clearpage
\bibliography{reference}

\newpage
\appendix

\section{Proof of Proposition \ref{proposition-1}}
\label{appendix:prop-decomposable}
\begin{proof} Let us first prove the implication \textsl{(1)}$\implies$\textsl{(2)}. For a decomposable MAMG $\mathcal{MG}(N, \boldsymbol{\mathcal{S}}, \boldsymbol{\mathcal{A}}, \mathbb{P}, R, \gamma, \pi_0)$, we have
\begin{equation*}
\begin{aligned}
\big[TQ\big](\vs,\va)&=R(\vs,\va)+\gamma\cdot\E_{\rvs'\sim\mathbb{P}}\big[ \max_{\va'}Q(\rvs', \va')\big]\nonumber\\
&= \sum_{i=1}^{N}R_i(s^i, a^i) + \gamma\int_{\boldsymbol{\mS}}\max_{\va'}Q(\rvs', \va') \left(\sum_{i=1}^{N}F_i(\rvs'|s^i, a^i)\right)d\vs'\\
&= \sum_{i=1}^{N}\left[R_i(s^i,a^i)+\gamma\int_{\boldsymbol{\mS}} \max_{\va'}Q(\rvs', \va')\cdot F_i(\rvs'|s^i, a^i)d\vs'\right] \triangleq \sum_{i=1}^{N} G^{(\gamma)}_i(s^i,a^i).
\end{aligned}
\end{equation*}
On the other hand, if statement \textsl{(2)} holds for any $\gamma$ and $\pi$, by setting $\gamma=0$ and expanding the Bellman operator $[TQ](\vs,\va)=R(\vs,\va)+\gamma\cdot\E_{\rvs'\sim\mathbb{P}}[ \max_{\va'}Q(\rvs', \va')]$, we obtain $\sum_{i=1}^{N}G^{(0)}_i(s^i, a^i) = R(\vs, \va)$. Hence, $R(\vs,\va)$ is a decomposable function, meaning that there exist functions $R_1, R_2,\ldots, R_N$ such that:
\[R(\vs,\va)=R_1(s^1,a^1)+R_2(s^2,a^2)+\ldots+R_N(s^N,a^N).\]
With this decomposition, for an arbitrary $\gamma>0$, we can rewrite the Bellman operator as
\begin{align}
    [TQ](\vs, \va) &= \sum_{i=1}^{N}G^{(\gamma)}_i(s^i, a^i) \nonumber\\
    &= \sum_{i=1}^{N}R_i(s^i, a^i) + \gamma\E_{\rvs'\sim\mathbb{P}}\big[ \max_{\boldsymbol{a}'}Q(\rvs', \boldsymbol{a}')\big] = \sum_{i=1}^{N}R_i(s^i, a^i) + \gamma\E_{\rvs'\sim\mathbb{P}}\big[ \max_{\boldsymbol{a}'}V^{\pi_Q}(\rvs')\big],\nonumber
\end{align}
where $\pi_Q$ is a greedy policy with respect to $Q$.
Let us set $g^{(\gamma)}_i(s^i, a^i) = G^{(\gamma)}_i(s^i, a^i) - R_i(s^i, a^i)$. The above equality implies that
\begin{align}
    \label{eq:decomposability-of-exp-v}
    \sum_{i=1}^{N} g^{(\gamma)}_i(s^i, a^i)
    = \gamma \E_{\rvs'\sim\mathbb{P}}\big[ V^{\pi_Q}(\rvs') \big] 
    = \gamma\langle \mathbb{P}(\cdot |\vs, \boldsymbol{a}), V^{\pi_Q}(\cdot) \rangle_{\mathcal{S}},
\end{align}
where $\langle \mathbb{P}(\cdot|\vs, \boldsymbol{a}), v(\cdot) \rangle_{\boldsymbol{\mathcal{S}}} = \int_{\boldsymbol{\mathcal{S}}} \mathbb{P}(s'|\vs, \boldsymbol{a})v(\vs') d\vs'$ is a linear functional of $v:\boldsymbol{\mathcal{S}}\rightarrow\mathbb{R}$. Hence, the decomposability of $\mathbb{P}(\cdot|\vs, \boldsymbol{a})$ follows from taking a functional derivative of Equation (\ref{eq:decomposability-of-exp-v}) with respect to $v(\cdot)$, which finishes the proof.
\end{proof}

\section{Proofs of results relating functions and their $N$-compositions}
\label{appendix:proofs-N-compositions}

\subsection{Proof of Lemma \ref{lemma:covering-n-sum}}

\begin{proof}
Let $\mathcal{M}^*$ be a minimal $\delta/N$-covering of $\mathcal{M}$. Let $m^{(N)}\in\mathcal{M}^{\oplus N}$. Then, there exist functions $m_1, \dots, m_N \in \mathcal{M}$, such that for any  $\boldsymbol{s} = s^{1:N}\in\boldsymbol{\mathcal{S}}$ and $\boldsymbol{a}=a^{1:N}\in\boldsymbol{\mathcal{A}}$, we have
\begin{align}
    m^{(N)}(\boldsymbol{s}, \boldsymbol{a}) = \sum\limits_{i=1}^{N}m_i(s^i, a^i).\nonumber
\end{align}
From the definition of a $\delta/N$-covering, we know that there exist $m_1^*, \dots, m_N^*$, such that for any $i\in[N]$, we have $||m_i - m_i^*||_{\infty}\leqslant \delta/N$. Hence, for any $\boldsymbol{s}=s^{1:N}$ and $\boldsymbol{a}=a^{1:N}$,
\begin{align}
    &\delta \geq \sum\limits_{i=1}^{N}|m_i(s^i, a^i) - m_i^*(s^i, a^i)|
    \geq \left| \sum\limits_{i=1}^{N}\left[m_i(s^i, a^i) - m_i^*(s^i, a^i)\right]\right| = \left| m^{(N)}(\boldsymbol{s}, \boldsymbol{a}) - \sum\limits_{i=1}^{N} m_i^*(s^i, a^i)\right|.\nonumber
\end{align}
As $\sum\limits_{i=1}^{N}m_i^*(\cdot, \cdot) \in \mathcal{M}^{\oplus N}$, it follows that $(\mathcal{M}^*)^{\oplus N}$ is a $\delta$-covering of $\mathcal{M}^{\oplus N}$.
We also have
\begin{align}
    \left| (\mathcal{M}^*)^{\oplus N} \right| \leqslant |\mathcal{M}^*|^N,\nonumber
\end{align}
which finishes the proof.
\end{proof}

\subsection{Proof of Lemma \ref{lemma:bounding-distance}}

\begin{proof}
Let $m_1^{(N)}\in\mathcal{M}_1^{\oplus N}$ and $m_2^{(N)}\in\mathcal{M}_2^{\oplus N}$.
For any $\boldsymbol{s}=s^{1:N}\in\boldsymbol{\mathcal{S}}, \ \boldsymbol{a}=a^{1:N}\in\boldsymbol{\mathcal{A}}$, we have
\begin{align}
    &\big|m_1^{(N)}(\boldsymbol{s}, \boldsymbol{a}) - m_2^{(N)}(\boldsymbol{s}, \boldsymbol{a})\big| = 
    \Big| \sum\limits_{i=1}^{N}m_{1, i}(s^i, a^i) - \sum\limits_{i=1}^{N}m_{2, i}(s^i, a^i) \Big| \nonumber\\
    & \quad \leqslant\sum\limits_{i=1}^{N}\big|m_{1, i}(s^i, a^i)- m_{2, i}(s^i, a^i)\big|
    \leqslant \sum\limits_{i=1}^{N} ||m_{1, i} - m_{2, i}||_{\infty}. \nonumber 
\end{align}
Therefore, taking supremum over $(\boldsymbol{s}, \boldsymbol{a})$, we have
\begin{align}
    \label{ineq:composition-components}
    ||m_1^{(N)} - m_2^{(N)}||_{\infty} \leqslant \sum_{i=1}^{N} ||m_{1, i} - m_{2, i}||_{\infty}. 
\end{align}
Let us now fix $m^{(N)}_1 = \widetilde{m}^{(N)}_1$. For every $i\in[N]$, let $\big( m_{2, i, k} \big)_{k\in\mathbb{N}}$ be a sequence in $\mathcal{M}_2$ such that
\begin{align}
    \label{eq:sequence-of-response-maps-infimum}
    \lim\limits_{k\rightarrow\infty}||\widetilde{m}_{1, i} - m_{2, i, k}||_{\infty} = \inf_{m_{2, i}\in\mathcal{M}_2} ||\widetilde{m}_{1, i} - m_{2, i}||_{\infty}.
\end{align}
The Inequality (\ref{ineq:composition-components}) implies that
\begin{align}
    \label{ineq:inequality-between-sequences}
    ||\widetilde{m}_1^{(N)} - m_{2, k}^{(N)}||_{\infty} \leqslant \sum_{i=1}^{N}||\widetilde{m}_{1, i} - m_{2, i, k}||_{\infty}.
\end{align}
As the right-hand side of the above inequality has a finite limit, given in Equation (\ref{eq:sequence-of-response-maps-infimum}), the sequence on the left-hand side above is bounded. Therefore, by Bolzano-Weierstrass Theorem, it has a convergent subsequence $\big( ||\widetilde{m}_1^{(N)} - m_{2, k_j}^{(N)}||_{\infty} \big)_{j\in\mathbb{N}}$. This and Inequality (\ref{ineq:inequality-between-sequences}) imply that
\begin{align}
    &\lim\limits_{j\rightarrow\infty}||\widetilde{m}_1^{(N)} - m_{2, k_j}^{(N)}||_{\infty} \leqslant \lim\limits_{j\rightarrow\infty}\sum_{i=1}^{N}||\widetilde{m}_{1, i} - m_{2, i, k_j}||_{\infty} \nonumber\\
    &= \sum_{i=1}^{N}\lim\limits_{j\rightarrow\infty}||\widetilde{m}_{1, i} - m_{2, i, k_j}||_{\infty} =
    \sum_{i=1}^{N}\inf_{m_{2, i}\in\mathcal{M}_2} ||\widetilde{m}_{1, i} - m_{2, i}||_{\infty}.\nonumber
\end{align}
We can therefore conclude that 
\begin{align}
    \label{ineq:distribution-of-infima}
    \inf_{m_2^{(N)}}||\widetilde{m}^{(N)}_1 - m^{(N)}_2||_{\infty} \leqslant \sum_{i=1}^{N}\inf_{m_{2, i}} ||\widetilde{m}_{1, i} - m_{2, i}||_{\infty}\\
    \text{(Here we dropped the sets from } inf \text{ for brevity.)}\nonumber
\end{align}
Now, unfreezing $\widetilde{m}^{(N)}_1$ and taking the supremum over $m_1^{(N)}\in\mathcal{M}_1^{\oplus N}$,
\begin{align}
    \sup_{m^{(N)}_1}\inf_{m_2^{(N)}}||m^{(N)}_1 - m^{(N)}_2||_{\infty} \leqslant \sup_{m^{(N)}_1}\sum_{i=1}^{N}\inf_{m_{2, i}} ||m_{1, i} - m_{2, i}||_{\infty} \leqslant 
    \sum_{i=1}^{N}\sup_{m_{1, i}}\inf_{m_{2, i}} ||m_{1, i} - m_{2, i}||_{\infty}.\nonumber
\end{align}
Recalling that suprema and infima over $m_{1, i}$ and $m_{2, i}$, for all $i\in[N]$, are taken over sets $\mathcal{M}_1$ and $\mathcal{M}_2$, respectively, allows us to rewrite the above as
\begin{align}
    \text{dist}\big( \mathcal{M}_1^{\oplus N}, \mathcal{M}_2^{\oplus N} \big) \leqslant N\cdot\text{dist}(\mathcal{M}_1, \mathcal{M}_2),\nonumber
\end{align}
which finishes the proof.
\end{proof}
\begin{Remark}
We would like to highlight that this result (Lemma \ref{lemma:bounding-distance}) is quite surprising. The presence of infimum in Definition \ref{def:sup-inf-distance} had thrown doubt on the possibility of decomposing the distance to $\mathcal{M}_2^{\oplus N}$ over the summing $N$ copies of $\mathcal{M}_2$, as it happened in Inequality (\ref{ineq:distribution-of-infima}). Indeed, for any collection of sets $\{\mathcal{X}_1, \dots, \mathcal{X}_N\}$, and any subset $\mathcal{Y}$ of $\mathcal{X}_1 \times \dots \times \mathcal{X}_N$, we have
\begin{align}
    \inf_{(x_1, \dots, x_N)\in \mathcal{Y}}\sum\limits_{i=1}^{N}x_i \geqslant  \sum\limits_{i=1}^{N}\inf_{x_i\in\mathcal{X}_i}x_i.
\end{align}
What enabled us to arrive there was the trick with a sequence of independent maps in $\mathcal{M}_2$ from Equation (\ref{eq:sequence-of-response-maps-infimum}), which always have a representant (composition map) in $\mathcal{M}_2^{\oplus N}$, for which they provide the upper bound from Inequality (\ref{ineq:inequality-between-sequences}).
\end{Remark}

\section{Proof of Theorem \ref{theorem:deep-vdn-convergence}}
\label{appendix:vdn-convergence}
\begin{proof}
Let us recall that by Theorems \ref{theorem:error-prop} \& \ref{theorem:one-step-approx}, we have
\begin{align}
    \label{ineq:write-out-error}
    &||Q^* - Q^{\pi_K}||_{1, \mu} \leqslant \frac{2\phi_{\mu, \sigma}\gamma}{(1-\gamma)^2}\cdot \epsilon_{\max} + \frac{4\gamma^{K+1}}{(1-\gamma)^2}R_{\max}\nonumber\\
    &\leqslant \frac{2\phi_{\mu, \sigma}\gamma}{(1-\gamma)^2}\cdot \left[
    4\text{dist}(\mathcal{F}^{\oplus N}_{\text{net}}, \mathcal{G}^{\oplus N}_{\text{H}})^2 + C\cdot(Q^2_{\max}/n)\cdot\log\mathcal{N}(\mathcal{F}^{\oplus N}_{\text{net}}, \delta, ||\cdot||_{\infty})
    \right]^{\frac{1}{2}} + \frac{4\gamma^{K+1}}{(1-\gamma)^2}R_{\max}.
\end{align}
Hence, to prove the theorem it remains to provide bounds for 
\begin{align}
    \text{dist}(\mathcal{F}^{\oplus N}_{\text{net}}, \mathcal{G}^{\oplus N}_{\text{H}}) \  \ \text{and} \ \ \log\mathcal{N}(\mathcal{F}^{\oplus N}_{\text{net}}, \delta, ||\cdot||_{\infty}).\nonumber
\end{align}
\textbf{Step 1 (Covering Numbers).} We begin with the latter. By Lemma \ref{lemma:covering-n-sum}, we have
\begin{align}
    \label{ineq:covering-local-global}
    \log\mathcal{N}(\mathcal{F}^{\oplus N}_{\text{net}}, \delta, ||\cdot||_{\infty}) \leqslant N \log \mathcal{N}(\mathcal{F}_{\text{net}}, \delta/N).
\end{align}
Furthermore, by Theorem 14.5 in \citep{anthony1999neural}, setting $D= \prod_{l=1}^{L^*+1}(d^*_l + 1)$, we have
\begin{align}
    \label{ineq:theorem-about-covering}
    \log\left[ \mathcal{N}\left(\frac{\delta}{N}, \mathcal{F}\left(L^*, \{d^*_j\}_{j=0}^{L^*+1}, s^*, \frac{Q_{\max}}{N}\right), ||\cdot||_{\infty} \right)\right] \leqslant (s^*+1)\cdot \log\left[ 2\frac{N}{\delta}\cdot (L^* + 1)\cdot D^2\right].
\end{align}
Let the covering number of the above be denoted as $\mathcal{N}_{\delta/N}$, so that the left-hand side equals $\log\mathcal{N}_{\delta/N}$. The class $\mathcal{F}_{\text{net}}$ consists of $|\mathcal{A}|$ components, each being a copy of the class  $\mathcal{F}\big(L^*, \{d^*_j\}_{j=0}^{L^*+1}, s^*, \frac{Q_{\text{max}}}{N} \big)$. Hence, by copying the $\delta/N$-covering of cardinality $\mathcal{N}_{\delta/N}$ to each of the class copies, we obtain a $\delta/N$-covering of $\mathcal{F}_{\text{net}}$ (by composing elements from all component classes). The resulting $\delta/N$-covering of $\mathcal{F}_{\text{net}}$ has $\mathcal{N}_{\delta/N}^{|\mathcal{A}|}$ elements. Hence
\begin{align}
    \mathcal{N}(\mathcal{F}_{\text{net}}, \delta/N) \leqslant \mathcal{N}_{\delta/N}^{|\mathcal{A}|}, \nonumber
\end{align}
which combined with Inequality (\ref{ineq:theorem-about-covering}), and with $\delta = \frac{1}{n}$ (Theorem \ref{theorem:one-step-approx} holds for any $\delta$) gives
\begin{align}
    \log\mathcal{N}(\mathcal{F}_{\text{net}}, \delta/N)
    \leqslant |\mathcal{A}|\cdot(s^*+1)\cdot \log\left[ 2N\cdot n \cdot (L^* + 1)\cdot D^2\right].
\end{align}
Furthermore, we have
\begin{align}
   &\log [ 2N\cdot n\cdot(L^*+1)\cdot D^2]
   =  \log [ 2 n\cdot(L^*+1)\cdot D^2] + \log(N) \nonumber\\
   &\leqslant \log [ 2 n\cdot(L^*+1)\cdot D^2] \left( 1+ \log(N) \right)
   \leqslant C_0\cdot \log [ 2 n\cdot(L^*+1)\cdot D^2] \cdot \log(N),\nonumber
\end{align}
where $C_0>0$ is an absolute constant. Recall the choice of hyper-parameters (Equation \ref{eq:hyperparams2}). We have
\begin{align}
     \log\mathcal{N}(\mathcal{F}_{\text{net}}, \delta/N) &\leqslant C_0\cdot |\mathcal{A}|\cdot (s^*+1)\cdot \log[2n\cdot (L^*+1)\cdot D^2]\cdot\log(N)\nonumber\\
     &= \mathcal{O}\left(|\mathcal{A}|\cdot s^* \cdot L^* \cdot\left(\log n + \max_{j\in[L^*]}\log(d^*_j)\right)\cdot\log(N) \right)\nonumber\\
     &= \mathcal{O}\left( |\mathcal{A}|\cdot n^{\alpha^*}(\log n)^{\xi^*} \cdot (\log n)^{\xi^*}(\log n + \xi^*\log n)\cdot\log(N)\right)\nonumber\\
     &= \mathcal{O}\left( |\mathcal{A}|\cdot n^{\alpha^*} \cdot (\log n)^{2\xi^*+1}\cdot\log(N)\right).\nonumber
\end{align}
Combining this with Inequality (\ref{ineq:covering-local-global}), we get that for some absolute constant $C_1>0$, 
\begin{align}
    \label{ineq:bound-on-covering}
    \log\mathcal{N}\left(\mathcal{F}^{\oplus N}_{\text{net}}, \frac{1}{n}, ||\cdot||_{\infty}\right) \leqslant 
    C_1\cdot N\cdot|\mathcal{A}|\cdot n^{\alpha^*} \cdot (\log n)^{2\xi^*+1}\cdot\log(N).
\end{align}
\textbf{Step 2 (Distance). } We now bound the distance
\begin{align}
    \text{dist}(\mathcal{F}^{\oplus N}_{\text{net}}, \mathcal{G}^{\oplus N}_{\text{H}}). \nonumber
\end{align}
By Lemma \ref{lemma:bounding-distance}, we have
\begin{align}
    \label{ineq:recall-distance-bound}
     \text{dist}(\mathcal{F}_{\text{net}}, \mathcal{G}^{\oplus N}_{\text{H}}) \leqslant N\cdot  \text{dist}(\mathcal{F}_{\text{net}}, \mathcal{G}_{\text{H}}),
\end{align}
which implies that it suffices to study the distance between the agents' local function classes. We invoke the following lemma.
\begin{Lemma}[Inequality 4.18, \citep{fan2020theoretical}]
For function classes $\mathcal{F}_{\text{net}}$  and $\mathcal{G}_{\text{H}}$ defined as in Definition \ref{definition:function-classes}, with hyper-parameters specified in Equations (\ref{eq:hyperparams1}) \& (\ref{eq:hyperparams2}),
\begin{align}
    \text{dist}\left(\mathcal{F}_{\text{net}}, \mathcal{G}_{\text{H}}\right)^2 = \mathcal{O}(n^{\alpha^*-1}).\nonumber
\end{align}
\end{Lemma}
Combining the lemma with Inequality (\ref{ineq:recall-distance-bound}), we obtain that for some absolute constant $C_2>0$, we have
\begin{align}
    \label{ineq:bound-on-distance}
    \text{dist}(\mathcal{F}^{\oplus N}_{\text{net}}, \mathcal{G}^{\oplus N}_{\text{H}})^2 \leqslant C_2\cdot N^2\cdot n^{\alpha^*-1}.
\end{align}
Combining Inequalities (\ref{ineq:bound-on-covering}) \& (\ref{ineq:bound-on-distance}) with Inequality (\ref{ineq:write-out-error}), we have
\begin{align}
    &||Q^* - Q^{\pi_K}||_{1, \mu} \leqslant \frac{2\phi_{\mu, \sigma}\gamma}{(1-\gamma)^2}\cdot \epsilon_{\max} + \frac{4\gamma^{K+1}}{(1-\gamma)^2}R_{\max}\nonumber\\
    &\leqslant \frac{2\phi_{\mu, \sigma}\gamma}{(1-\gamma)^2}\cdot \Big[
    4C_2 \cdot N^2 \cdot n^{\alpha^*-1} \nonumber\\
    &\quad \quad \quad + C\cdot(Q^2_{\max}/n)\cdot 
     C_1\cdot N\cdot|\mathcal{A}|\cdot n^{\alpha^*}\cdot (\log n)^{2\xi^*+1} \cdot\log(N) \Big]^{\frac{1}{2}} + \frac{4\gamma^{K+1}}{(1-\gamma)^2}R_{\max}\nonumber\\
    &\leqslant \frac{2\phi_{\mu, \sigma}\gamma}{(1-\gamma)^2}\cdot \Big[
    2\sqrt{C_2} \cdot N \cdot n^{(\alpha^*-1)/2} \nonumber\\
    &\quad \quad \quad + \sqrt{C\cdot C_1\cdot N\cdot\log(N)\cdot|\mathcal{A}| }\cdot Q_{\max}\cdot n^{(\alpha^*-1)/2} \cdot (\log n)^{\xi^*+1/2}  \Big] + \frac{4\gamma^{K+1}}{(1-\gamma)^2}R_{\max},\nonumber\\
    &\text{simplifying, taking into account that }1-\alpha^*>0,\text{ and bounding,}\nonumber\\
    &\leqslant  \frac{\widetilde{C}\phi_{\mu, \sigma}\gamma}{(1-\gamma)^2}\cdot \Big[
    N \cdot n^{-(1 - \alpha^*)/2} \nonumber\\
    &\quad \quad \quad + \sqrt{ N\cdot\log(N)\cdot|\mathcal{A}| }\cdot Q_{\max}\cdot n^{-(1-\alpha^*)/2} \cdot (\log n)^{\xi^*+1/2}  \Big] + \frac{4\gamma^{K+1}}{(1-\gamma)^2}R_{\max},\nonumber
\end{align}
where $\widetilde{C}>0$ is an absolute constant. This completes the proof.
\end{proof}

\section{Proofs of Lemmas and Theorems in Section \ref{sec-non}}
\label{appendix-d}
In this section, we propose the complete proofs for the lemmas and theorems in Section \ref{sec-non}. 
\subsection{Proof of Lemma \ref{lemma-non-1}}
According to the Bellman Equation, we can obtain that:
\begin{equation*}
\begin{aligned}
Q^*(s,a)&=R(s,a)+\gamma\mathbb{E}_{s'|s,a}V^*(s'):=R(s,a)+\gamma P^{\pi^*}Q^*(s,a)\\
Q^{\pi^k}(s,a)&= R(s,a)+\gamma\mathbb{E}_{s'|s,a}V^{\pi_k}(s'):=R(s,a)+\gamma P^{\pi_k}Q^{\pi_k}(s,a)
\end{aligned}
\end{equation*}
Therefore, we subtract the first equation with the second, and know that for $\forall(s,a)\in\mS\times\mA$:
\begin{equation}
\label{eqn-proof-non-1-1}
\begin{aligned}
&\left(Q^*-Q^{\pi_k}\right)(s,a)=\gamma\cdot\left(P^{\pi^*}Q^*(s,a)-P^{\pi_k}Q^{\pi_k}(s,a)\right)\\
\overset{(a)}{\leqslant} ~& \gamma\cdot\left(P^{\pi^*}Q^*-P^{\pi_k}Q^*+P^{\pi_k}Q^*-P^{\pi_k}Q^{\pi_k}-P^{\pi^*}\widetilde{Q}_k+P^{\pi_k}\widetilde{Q}_k\right)(s,a)\\
=~&\gamma\cdot\left(P^{\pi^*}-P^{\pi_k}\right)\left(Q^*-\widetilde{Q}_k\right)(s,a)+\gamma\cdot P^{\pi_k}\left(Q^*-Q^{\pi_k}\right)(s,a)\\
\overset{(b)}{\leqslant}~& 2\gamma\cdot\|Q^*-\widetilde{Q}_k\|_{\infty}+\gamma\cdot\|Q^*-Q^{\pi_k}\|_{\infty}.
\end{aligned}    
\end{equation}
Since the $(s,a)\in\mS\times\mA$ can be chosen arbitrarily, so we conclude that:
\[\|Q^*-Q^{\pi_k}\|_{\infty}\leqslant 2\gamma\cdot\|Q^*-\widetilde{Q}_k\|_{\infty}+\gamma\cdot\|Q^*-Q^{\pi_k}\|_{\infty}\Rightarrow\|Q^*-Q^{\pi_k}\|_{\infty}\leqslant \frac{2\gamma}{1-\gamma}\|Q^*-\widetilde{Q}_k\|_{\infty},\]
which comes to our conclusion. In Equation (\ref{eqn-proof-non-1-1}), (a) and (b) hold because of two properties of the operator $T^{\pi}$, and we list them below as two lemmas. The first lemma shows us that for a given action-value function $Q:\mS\times\mA\rightarrow\mathbb{R}$, the policy $\pi$ that maximizes $T^{\pi}Q$ is exactly the greedy policy for $Q$, i.e., $\pi_Q$. 

\begin{lemma}
\label{lemma-non-1-1}
For any action value function $Q:\mS\times\mA\rightarrow\mathbb{R}$ and any policy $\pi$, denote $\pi_Q$ as the greedy policy for $Q$, then we have:
\[P^{\pi_Q}Q=PQ\geqslant P^{\pi}Q.\]
\end{lemma}

\begin{proof}[Proof of Lemma \ref{lemma-non-1-1}]
By the definition of the $P^{\pi}$ operator, we know that:
\[P^{\pi}Q(s,a)=\mathbb{E}_{s'|s,a}\mathbb{E}_{a'\sim\pi(s')}Q(s',a')\leqslant \mathbb{E}_{s'|s,a}\max_{a'}Q(s',a')=PQ(s,a)\]
holds for $\forall (s,a)\in\mS\times\mA$. Therefore, we can conclude that $PQ\geqslant P^{\pi}Q$ holds for any policy $\pi$ and action value function $Q$. On the other hand, since $\pi_Q$ is the greedy policy with regard to $Q$, we have:
\[\mathbb{P}\left[a\in\arg\max_{a'}Q(s,a')~\Big|a\sim\pi_Q(s)\right]=1,\]
which leads to 
\[P^{\pi_Q}Q(s,a)=\mathbb{E}_{s'|s,a}\mathbb{E}_{a'\sim\pi_Q(s')}Q(s',a')=\mathbb{E}_{s'|s,a}\max_{a'}Q(s',a')=PQ(s,a).\]
Combine the two equations above, and it comes to our conclusion. 
\end{proof}
The second lemma shows us that for any policy $\pi$, the operator $P^{\pi}$ has Lipschitz constant 1 under the $l_{\infty}$ norm. 
\begin{lemma}
\label{lemma-non-1-2}
For any policy $\pi$ and any two action value functions $Q_1, Q_2:\mS\times\mA\rightarrow \mathbb{R}$, we have:
\[\|P^{\pi}Q_1-P^{\pi}Q_2\|_{\infty}\leqslant \|Q_1-Q_2\|_{\infty}.\]
\end{lemma}
\begin{proof}[Proof of Lemma \ref{lemma-non-1-2}]
Since 
\[P^{\pi}Q_1(s,a)=\mathbb{E}_{s'|s,a}\mathbb{E}_{a'\sim\pi(s')}Q_1(s',a'), P^{\pi}Q_2(s,a)=\mathbb{E}_{s'|s,a}\mathbb{E}_{a'\sim\pi(s')}Q_2(s',a'),\]
we have:
\begin{equation*}
\begin{aligned}
&|P^{\pi}Q_1(s,a)-P^{\pi}Q_2(s,a)|\leqslant \mathbb{E}_{s'|s,a}\mathbb{E}_{a'\sim\pi(s')}|Q_1(s',a')-Q_2(s',a')|\\
\leqslant ~&\mathbb{E}_{s'|s,a}\mathbb{E}_{a'\sim\pi(s')}\|Q_1-Q_2\|_{\infty}=\|Q_1-Q_2\|_{\infty}.
\end{aligned}    
\end{equation*}
Since the state-action pair $(s,a)$ can be arbitrarily chosen, we obtain that:
\[\|P^{\pi}Q_1-P^{\pi}Q_2\|_{\infty}\leqslant \|Q_1-Q_2\|_{\infty},\]
which comes to our conclusion. 
\end{proof}
Since $\pi_k$ is the greedy policy with regard to $\widetilde{Q}_k$, we know that $P^{\pi_k}\widetilde{Q}_k\geqslant P^{\pi^*}\widetilde{Q}_k$ by Lemma \ref{lemma-non-1-1}, which explains why (a) holds. Also, (b) is a direct extension of Lemma \ref{lemma-non-1-2}.

\subsection{Proof of Lemma \ref{lemma-non-2}}
As we know that, the estimations of optimal action value function $Q^*$ are iteratively updated by:
\[\widetilde{Q}_{k+1}=\arg\min_{f\in\mathcal F_{k+1}}\frac{1}{n}\sum_{i=1}^{n}\left[T\widetilde{Q}_k(s_i,a_i)-f(s_i,a_i)\right]^2,\]
where $TQ(s,a)=R(s,a)+\gamma\cdot PQ(s,a)$. Also, we denote When $n$ is sufficiently large and $\mathcal F$ is closed in the set of decomposable continuous functions $\mathcal C^{\oplus N}$, we know that 
\[\widetilde{Q}_{k+1}\approx \arg\min_{f\in\mathcal C^{\oplus N}}\mathbb{E}_{(s,a)\sim\sigma}\left[T\widetilde{Q}_k(s,a)-f(s,a)\right]^2 := \Proj\left(T\widetilde{Q}_k, \mathcal C^{\oplus N}, \|\cdot\|_{\sigma}\right).\]
After we denote 
\[\varepsilon_{\max}=\max_{k\in [K]}\left\|\widetilde{Q}_{k+1}-\Proj\left(T\widetilde{Q}_k, \mathcal C^{\oplus N}, \|\cdot\|_{\sigma}\right)\right\|_{\infty},\]
we have:
\begin{equation}
\label{eqn-proof-non-2-1}    
\begin{aligned}
\|Q^*-\widetilde{Q}_{k+1}\|_{\infty}&\leqslant \left\|\widetilde{Q}_{k+1}-\Proj\left(T\widetilde{Q}_k, \mathcal C^{\oplus N}, \|\cdot\|_{\sigma}\right)\right\|_{\infty}+\left\|Q^*-\Proj\left(T\widetilde{Q}_k, \mathcal C^{\oplus N}, \|\cdot\|_{\sigma}\right)\right\|_{\infty}\\
&\leqslant \varepsilon_{\max}+\left\|\Proj\left(Q^*,\mathcal C^{\oplus N}, \|\cdot\|_{\sigma}\right)-\Proj\left(T\widetilde{Q}_k, \mathcal C^{\oplus N}, \|\cdot\|_{\sigma}\right)\right\|_{\infty}\\
&\overset{(a)}{\leqslant} \varepsilon_{\max}+ (2N-1)\cdot\|Q^*-T\widetilde{Q}_k\|_{\infty}\overset{(b)}{\leqslant} \varepsilon_{\max}+ (2N-1)\gamma\cdot\|Q^*-\widetilde{Q}_k\|_{\infty}
\end{aligned}
\end{equation}
Here, (b) holds because by using Lemma \ref{lemma-non-1-2}:
\[\|Q^*-T\widetilde{Q}_k\|_{\infty}=\|TQ^*-T\widetilde{Q}_k\|_{\infty}=\gamma\cdot\|PQ^*-P\widetilde{Q}_k\|_{\infty}\leqslant \gamma\cdot\|Q^*-\widetilde{Q}_k\|_{\infty}.\]
(a) holds because of the Lipschitz property for the projection operator, and we are going to explain this in the following lemma, and meanwhile we will give an explicit form for the projection operator. 
\begin{lemma}[Explicit form of Projection Operator]
\label{lemma-proj}
For the projection operator above, we have the explicit expression when distribution $\sigma$ is separable, which means $\sigma\in\mathcal P(\mS\times \mA)$ can be written as $\sigma_1\times\sigma_2\times\ldots\times\sigma_N$ where $\sigma_i\in\mathcal P(\Upsilon\times \mathcal{A}^{(i)})$ is a distribution over the subspace. Actually, for any $C^1$ continuous function $f:[a,b]^N\rightarrow \mathbb{R}$, the closet decomposable $C^1$ continuous function is: 
\[\Proj\left(f, \mathcal C^{\oplus N}, \|\cdot\|_{\sigma}\right)(x_1,x_2,\ldots,x_N) = \sum_{i=1}^{N}f_i(x_i)-(N-1)C.\]
where $f_i(x_i)=\mathbb{E}_{x_{-i}\sim\sigma_{-i}}[f(x_i, x_{-i})],~\forall i\in[N]$ and $C=\mathbb{E}_{x\sim\sigma}f(x)$. 
\end{lemma}

\begin{proof}[Proof of Lemma \ref{lemma-proj}] 
For brevity, we denote $\sigma_i$ as the marginal distribution of $x_i$, and $\sigma_{-i}$ as the marginal distribution of $x_{-i}\in\mathbb{R}^{N-1}$. We have
\begin{align}
&\E_{x\sim\sigma}\left[ \left( \sum_{i=1}^{N}f_i(x_i) - f(x) \right)^2\right] = \E_{x_i\sim\sigma_i}\left[ 
    \E_{x_{-i}\sim\sigma_{-i}}\left[\left( \sum_{i=1}^{N}f_i(x_i) - f(x) \right)^2\right]  \right]\nonumber\\
    &= \E_{x_i\sim\sigma_i}\left[ 
    \E_{x_{-i}\sim\sigma_{-i}}\left[ \left(f_i(x_i) + \left( \sum_{j\neq i}f_j(x_j) - f(x) \right) \right)^2\right]  \right]\nonumber\\
    &= \E_{x_i\sim\sigma_i}\left[ f_i(x_i)^2 + 2f_i(x_i)\E_{x_{-i}\sim\sigma_{-i}}\left[  \sum_{j\neq i}f_j(x_j) - f(x)  \right]
    + \E_{x_{-i}\sim\sigma_{-i}}\left[  \left(\sum_{j\neq i}f_j(x_j) - f(x) \right)^2 \right]
   \right].\nonumber
\end{align}
The minimum is thus attained if, for every $i$,
\begin{align}
    f_i(x_i) = \E_{x_{-i}\sim\sigma_{-i}}\left[ f(x_i, x_{-i}) - \sum_{j\neq i}f_j(x_j)  \right].\nonumber
\end{align}
Denoting $c_i := \E_{x_i\sim\sigma_i}[f_i(x_i)]$, then we have:
\begin{align}
    \label{eq:form-of-fi}
    f_i(x_i) = \E_{x_{-i}\sim\sigma_{-i}}\left[ f(x_i, x_{-i}) \right] - \sum_{j\neq i}c_j.
\end{align}
Taking expectation under $x_i\sim\sigma_i$ on both sides,
\begin{align}
    c_i = \E_{x\sim\sigma}[f(x)] - \sum_{j\neq i}c_j,\notag
\end{align}
which leads to 
\begin{align}
    C := \sum_{j=1}^{N}c_j = \E_{x\sim\sigma}[f(x)].
\end{align}
Combining this with equation \ref{eq:form-of-fi} and aggregating constants, we conclude that the closest decomposable function under distribution $\sigma$ is
\begin{align}
\sum_{i=1}^{N}f_i(x_i) - (N-1)C, \notag
\end{align}
where 
$f_i(x^i) = E_{x_{-i}\sim\sigma_{-i}}\left[ f(x_i, x_{-i}) \right]$ and $C= \E_{x\sim\sigma}[f(x)]$, which comes to our conclusion. 
\end{proof}

From this lemma, we can obtain two properties of the projection operator. First, for two functions $f,g:\mathbb{R}^N\rightarrow\mathbb{R}$, we know that \[\left\|\Proj\left(f,\mathcal C^{\oplus N}, \|\cdot\|_{\sigma}\right)-\Proj\left(g,\mathcal C^{\oplus N}, \|\cdot\|_{\sigma}\right)\right\|_{\infty}\leqslant (2N-1)\cdot\|f-g\|_{\infty}.\]
Second, if function $f$ has Lipschitz constant $L$, then its projection $\Proj\left(f,\mathcal C^{\oplus N}, \|\cdot\|_{\sigma}\right)$ is also Lipschitz continuous, and its Lipschitz constant is $\sqrt{N}L$ since for $\forall i\in[N]$:
\[|f_i(x_i)-f_i(x_i')|=\left|\mathbb{E}_{x_{-i}\sim\sigma_{-i}}[f(x_i, x_{-i})-f(x_i',x_{-i})]\right|\leqslant L\cdot |x_i-x_i'|.\]
After taking $i=1,2,\ldots, N$ and summing them up:
\[\left|\Proj\left(f,\mathcal C^{\oplus N}, \|\cdot\|_{\sigma}\right)(x)-\Proj\left(f,\mathcal C^{\oplus N}, \|\cdot\|_{\sigma}\right)(x')\right|\leqslant L\|x-x'\|_1 \leqslant L\sqrt{N}\|x-x'\|_2.\]

\subsection{Proof of Lemma \ref{lemma-non-3}}
For each iteration $k\in[K]$, we are going to upper bound the discrepancy of the $k$-th iteration
\[\varepsilon_k := \left\|\widetilde{Q}_{k+1}-\Proj\left(T\widetilde{Q}_k, \mathcal C^{\oplus N}, \|\cdot\|_{\sigma}\right)\right\|_{\infty}. \]
We know that:
\[\widetilde{Q}_{k+1}=\arg\min_{f\in\mathcal F}\frac{1}{n}\sum_{i=1}^{n}\left[f(s_i,a_i)-T\widetilde{Q}_k(s_i,a_i)\right]^2.\]
If we make our sample size $n$ large enough, the $\widetilde{Q}_{k+1}$ be closer to:
\[\arg\min_{f\in\mathcal F}\mathbb{E}_{(s,a)\sim\sigma}\left[f(s,a)-T\widetilde{Q}_k(s,a)\right]^2.\]

By using the posterior generalization bound proposed by Theorem \ref{thm-posterior}, we know that: for $\forall \ba\in\mathcal{A}$ and any distribution $\sigma_s$ over state space $\mathcal{S}$, with probability at least $1-\delta$ over the choice of training data, it holds that:
\begin{equation*}
\begin{aligned}
&\left|\left\|\widetilde{Q}_{k+1}(\cdot, \ba)-T\widetilde{Q}_k(\cdot,\ba)\right\|_n^2-\left\|\widetilde{Q}_{k+1}(\cdot, \ba)-T\widetilde{Q}_k(\cdot,\ba)\right\|_{\sigma_s}^2\right|\leqslant\\ &~~~~~~~~~~~~16Q_{\max}(\|\widetilde{Q}_{k+1}(\cdot,\ba)\|_{P}+1)\sqrt{\frac{2\log(2d)}{n}}+4Q_{\max}^2\sqrt{\frac{2\log(2c(\|\widetilde{Q}_{k+1}(\cdot,\ba)\|_{P}+1)^2/\delta)}{n}}
\end{aligned}    
\end{equation*}
From the inequality above, we try to establish an upper bound of $\left|\|\widetilde{Q}_{k+1}-T\widetilde{Q}_k\|_n^2-\|\widetilde{Q}_{k+1}-T\widetilde{Q}_k\|_{\sigma}^2\right|$ where $\sigma$ is a distribution over $\mS\times \mA$. According to our assumption, action space $\mA$ is a discrete space. Denote $\mA=\{\ba_1,\ba_2,\ldots,\ba_{|\mA|}\}$, then with probability at least $1-|\mA|\delta$:
\begin{align}
\label{eqn-lemma-non-3-1}
&\left|\|\widetilde{Q}_{k+1}-T\widetilde{Q}_k\|_n^2-\|\widetilde{Q}_{k+1}-T\widetilde{Q}_k\|_{\sigma}^2\right|\notag\\ =~& \left|\sum_{\ba\in\mA} \hat{p}_{\ba}\cdot\|\widetilde{Q}_{k+1}(\cdot,\ba)-T\widetilde{Q}_k(\cdot,\ba)\|_{n\hat{p}_{\ba}}^2-\sum_{\ba\in\mA} p_{\ba}\cdot \|\widetilde{Q}_{k+1}(\cdot,\ba)-T\widetilde{Q}_k(\cdot,\ba)\|_{\sigma_{\ba}}^2\right|\notag\\
\leqslant~& \sum_{\ba\in\mA}\hat{p}_{\ba}\cdot\left|\|\widetilde{Q}_{k+1}(\cdot,\ba)-T\widetilde{Q}_k(\cdot,\ba)\|_{n\hat{p}_{\ba}}^2-\|\widetilde{Q}_{k+1}(\cdot,\ba)-T\widetilde{Q}_k(\cdot,\ba)\|_{\sigma_{\ba}}^2\right|\notag\\
&~+\sum_{\ba\in\mA}|\hat{p}_{\ba}-p_{\ba}|\cdot \|\widetilde{Q}_{k+1}(\cdot,\ba)-T\widetilde{Q}_k(\cdot,\ba)\|_{\sigma_{\ba}}^2\notag\\
\leqslant~& \sum_{\ba\in\mA}\hat{p}_{\ba}\cdot\left(16Q_{\max}(\|\widetilde{Q}_{k+1}(\cdot,\ba)\|_{P}+1)\sqrt{\frac{2\log(2d)}{n\hat{p}_{\ba}}}+4Q_{\max}^2\sqrt{\frac{2\log(2c(\|\widetilde{Q}_{k+1}(\cdot,\ba)\|_{P}+1)^2/\delta)}{n\hat{p}_{\ba}}}\right)\notag\\
&~+\sum_{\ba\in\mA}|\hat{p}_{\ba}-p_{\ba}|\cdot \|\widetilde{Q}_{k+1}(\cdot,\ba)-T\widetilde{Q}_k(\cdot,\ba)\|_{\sigma_{\ba}}^2
\end{align}    

Here, for $\forall \ba'\in\mA$, $p_{\ba'}:=\mathbb{P}_{(s,\ba)\sim\sigma}[\ba=\ba']$ and $\hat{p}_{\ba'}:=\frac{1}{n}\sum_{i=1}^{n}\mathbb{I}\{\ba^i=\ba'\}$ stand for the population probability and the empirical probability of joint action $\ba'$ under distribution $\sigma\in\mathcal P(\mS\times\mA)$. By using Hoeffding inequality, we know that: for $\forall\ba\in\mA$, 
\[\mathbb{P}\left[|\hat{p}_{\ba}-p_{\ba}|>t\right]\leqslant 2\exp(-2nt^2),\]
which means with probability at least $1-\delta$, it holds that $|\hat{p}_{\ba}-p_{\ba}|\leqslant \sqrt{\frac{\log(2/\delta)}{n}}$. To sum up, with probability at least $1-2|\mA|\delta$, we have:
\begin{align}
\label{eqn-lemma-non-3-2}
&\left|\|\widetilde{Q}_{k+1}-T\widetilde{Q}_k\|_n^2-\|\widetilde{Q}_{k+1}-T\widetilde{Q}_k\|_{\sigma}^2\right|\notag\\
\leqslant~& \sum_{\ba\in\mA}\sqrt{\hat{p}_{\ba}}\cdot\left(16Q_{\max}(\|\widetilde{Q}_{k+1}(\cdot,\ba)\|_{P}+1)\sqrt{\frac{2\log(2d)}{n}}+4Q_{\max}^2\sqrt{\frac{2\log(2c(\|\widetilde{Q}_{k+1}(\cdot,\ba)\|_{P}+1)^2/\delta)}{n}}\right)\notag\\
&~+\sum_{\ba\in\mA}\sqrt{\frac{\log(2/\delta)}{n}}\cdot 4Q_{\max}^2.
\end{align}    

According to the way to construct $\widetilde{Q}_{k+1}$, we know that: $\|\widetilde{Q}_{k+1}(\cdot,\ba)\|_P\leqslant B_{k+1}~\forall\ba\in\mA$. Therefore, from Equation (\ref{eqn-lemma-non-3-2}), we obtain that:
\begin{align}
\label{eqn-lemma-non-3-3}
&\left|\|\widetilde{Q}_{k+1}-T\widetilde{Q}_k\|_n^2-\|\widetilde{Q}_{k+1}-T\widetilde{Q}_k\|_{\sigma}^2\right|\notag\\
\leqslant~& \sqrt{|\mA|}\cdot\left(16Q_{\max}(B_{k+1}+1)\sqrt{\frac{2\log(2d)}{n}}+4Q_{\max}^2\sqrt{\frac{2\log(2c(B_{k+1}+1)^2/\delta)}{n}}\right)\notag\\
+~&|\mA|\cdot\sqrt{\frac{\log(2/\delta)}{n}}\cdot 4Q_{\max}^2 := \Delta_1
\end{align}  
holds with probability at least $1-2|\mA|\delta$ over the sampling. Next, we are going to conduct the same upper bound for function $\widehat{Q}_{k+1}$. Denote the decomposable function:
\[\Proj\left(T\widetilde{Q}_k, \mathcal{C}^{\oplus N}, \|\cdot\|_{\sigma}\right) = f_k^1(\gamma_1,a_1)+f_k^2(\gamma_2,a_2)+\ldots+f_k^N(\gamma_N,a_N).\]
According to Theorem \ref{approx}, we know that for $\forall i\in[N], a_i\in\mathcal{A}^{(i)}$, there exists a two-layer network $\widehat{f}^i:\Upsilon^{(i)}\times\mathcal{A}^{(i)}\rightarrow \mathbb{R}$ of width $M$ such that $\|\widehat{f}^i(\cdot,a_i)-f_k^i(\cdot,a_i)\|_{P}\leqslant 4\gamma\left(f_k^i(\cdot, a_i)\right)$ and for any distribution $\sigma_{\gamma}$:
\[\mathbb{E}_{\gamma_i\sim \sigma_{\gamma}}\left(\widehat{f}^i(\gamma_i,a_i)-f_k^i(\gamma_i,a_i)\right)^2\leqslant \frac{16\gamma^2\left(f_k^i(\cdot,a_i)\right)}{M}.\]
Then, for any joint action $\ba=(a_1,a_2,\ldots,a_N)$, there exists a function $\widehat{Q}_{k+1}(s,\ba):= \widehat{f}^1(\gamma_1,a_1)+\widehat{f}^2(\gamma_2,a_2)+\ldots+\widehat{f}^N(\gamma_N,a_N)$, which satisfies the following two properties:
\begin{itemize}
\item Function $\widehat{Q}_{k+1}(\cdot, \ba) $ is a two-layer ReLU network with width $M|\mA|$ and its path norm \[\|\widehat{Q}_{k+1}(\cdot, \ba)\|_P\leqslant 4\sum_{i=1}^{N} \gamma(f_k^i(\cdot,a_i)).\]
In order to make $\widehat{Q}_{k+1}$ contained in the function class $\mathcal F(B_{k+1})^{\oplus N}$, we have to make:
\[B_{k+1} \geqslant 4N\cdot \max_{i,a_i}\gamma\left(f_k^i(\cdot, a_i)\right),\]
so that we can guarantee that $\widehat{Q}_{k+1}(\cdot,\ba)\in \mathcal F(B_{k+1})^{\oplus N}$.
\item For any distribution $\sigma_s$ over the state space $\mS$, the mean squared error can be upper bounded as:
\begin{equation}
\label{eqn-lemma-non-3-4}   
\begin{aligned}
&\mathbb{E}_{s\sim\sigma_s}\left(\widehat{Q}_{k+1}(s,\ba)-\Proj\left(T\widetilde{Q}_k, \mathcal{C}^{\oplus N}, \|\cdot\|_{\sigma}\right)\right)^2\leqslant N\cdot\sum_{i=1}^{N}\mathbb{E}_{s\sim\sigma_s}\left(\widehat{f}^i(\gamma_i,a_i)-f_k^i(\gamma_i,a_i)\right)^2\\
&~~\leqslant \frac{16N^2}{M}\cdot\max_{i,a_i}\gamma^2(f_k^i(\cdot,a_i))\leqslant \frac{B_{k+1}^2}{M}.
\end{aligned}
\end{equation}
\end{itemize}
Again, by using the same technique as Equation (\ref{eqn-lemma-non-3-1}), we know that 
\begin{align}
\label{eqn-lemma-non-3-5}
&\left|\|\widehat{Q}_{k+1}-T\widetilde{Q}_k\|_n^2-\|\widetilde{Q}_{k+1}-T\widetilde{Q}_k\|_{\sigma}^2\right|\notag\\
\leqslant~& \sum_{\ba\in\mA}\sqrt{\hat{p}_{\ba}}\cdot\left(16Q_{\max}(\|\widehat{Q}_{k+1}(\cdot,\ba)\|_P+1)\sqrt{\frac{2\log(2d)}{n}}+4Q_{\max}^2\sqrt{\frac{2\log(2c\|\widehat{Q}_{k+1}\|)}{n}}\right)\notag\\
&~+\sum_{\ba\in\mA}\sqrt{\frac{\log(2/\delta)}{n}}\cdot 4Q_{\max}^2
\end{align}
holds with probability at least $1-2|\mA|\delta$. Therefore, according to the two properties above, we can conclude that: with probability at least $1-2|\mA|\delta$, it holds that
\begin{align}
\label{eqn-lemma-non-3-6}
&\left|\|\widehat{Q}_{k+1}-T\widetilde{Q}_k\|_n^2-\|\widehat{Q}_{k+1}-T\widetilde{Q}_k\|_{\sigma}^2\right|\notag\\
\leqslant~& \sqrt{|\mA|}\cdot\left(16Q_{\max}(C_k+1)\sqrt{\frac{2\log(2d)}{n}}+4Q_{\max}^2\sqrt{\frac{2\log(2c(C_k+1)^2/\delta)}{n}}\right)\notag\\
&~~~+|\mA|\cdot\sqrt{\frac{\log(2/\delta)}{n}}\cdot 4Q_{\max}^2\notag\\
\leqslant~& \sqrt{|\mA|}\cdot\left(16Q_{\max}(B_{k+1}+1)\sqrt{\frac{2\log(2d)}{n}}+4Q_{\max}^2\sqrt{\frac{2\log(2c(B_{k+1}+1)^2/\delta)}{n}}\right)\notag\\
&~~~+|\mA|\cdot\sqrt{\frac{\log(2/\delta)}{n}}\cdot 4Q_{\max}^2 := \Delta_2
\end{align}  
where $C_k = 4\sum_{i=1}^{N}\gamma(f_k^i(\cdot,a_i))\leqslant B_{k+1}.$ After summing up Equation (\ref{eqn-lemma-non-3-3}) and Equation (\ref{eqn-lemma-non-3-6}), we know that with probability at least $1-4|\mA|\delta$ over sampling, the following two inequalities hold simultaneously:
\begin{equation*}
\begin{aligned}
&\left|\|\widetilde{Q}_{k+1}-T\widetilde{Q}_k\|_n^2-\|\widetilde{Q}_{k+1}-T\widetilde{Q}_k\|_{\sigma}^2\right|\leqslant |\mA|\cdot\sqrt{\frac{\log(2/\delta)}{n}}\cdot 4Q_{\max}^2\\
&+\sqrt{|\mA|}\cdot\left(16Q_{\max}(B_{k+1}+1)\sqrt{\frac{2\log(2d)}{n}}+4Q_{\max}^2\sqrt{\frac{2\log(2c(B_{k+1}+1)^2/\delta)}{n}}\right) := \Delta_1,\\
&\left|\|\widehat{Q}_{k+1}-T\widetilde{Q}_k\|_n^2-\|\widehat{Q}_{k+1}-T\widetilde{Q}_k\|_{\sigma}^2\right|\leqslant |\mA|\cdot\sqrt{\frac{\log(2/\delta)}{n}}\cdot 4Q_{\max}^2\\
&+ \sqrt{|\mA|}\cdot\left(16Q_{\max}(B_{k+1}+1)\sqrt{\frac{2\log(2d)}{n}}+4Q_{\max}^2\sqrt{\frac{2\log(2c(B_{k+1}+1)^2/\delta)}{n}}\right):= \Delta_2
\end{aligned}    
\end{equation*}
Then: under the events above, by the definition of $\widetilde{Q}_{k+1}$, we have:
\[\|\widetilde{Q}_{k+1}-T\widetilde{Q}_k\|_{\sigma}^2\leqslant \|\widetilde{Q}_{k+1}-T\widetilde{Q}_k\|_n^2 + \Delta_1 \leqslant \|\widehat{Q}_{k+1}-T\widetilde{Q}_k\|_n^2 + \Delta_1\leqslant \|\widehat{Q}_{k+1}-T\widetilde{Q}_k\|_{\sigma}^2 + \Delta_1+\Delta_2. \]
Note that for any decomposable continuous function $f\in\mathcal C^{\oplus N}$:
\[\|f-T\widetilde{Q}_k\|_{\sigma}^2=\left\|f-\Proj\left(T\widetilde{Q}_k, \mathcal C^{\oplus N}, \|\cdot\|_{\sigma}\right)\right\|_{\sigma}^2+\left\|\mathrm{Proj}\left(T\widetilde{Q}_k, \mathcal C^{\oplus N}, \|\cdot\|_{\sigma}\right)-T\widetilde{Q}_k\right\|_{\sigma}^2.\]
Therefore, we conclude that:
\begin{equation}
\label{eqn-lemma-non-3-7}
\left\|\widetilde{Q}_{k+1}-\mathrm{Proj}\left(T\widetilde{Q}_k, \mathcal C^{\oplus N}, \|\cdot\|_{\sigma}\right)\right\|_{\sigma}^2\leqslant \left\|\widehat{Q}_{k+1}-\mathrm{Proj}\left(T\widetilde{Q}_k, \mathcal C^{\oplus N}, \|\cdot\|_{\sigma}\right)\right\|_{\sigma}^2 + \Delta_1+\Delta_2.
\end{equation}

We have already obtained upper bound for all the three terms. After adding them up, we obtain the following bound:
\begin{equation}
\label{eqn-lemma-non-3-8}
\begin{aligned}
&\left\|\widetilde{Q}_{k+1}-\mathrm{Proj}\left(T\widetilde{Q}_k, \mathcal C^{\oplus N}, \|\cdot\|_{\sigma}\right)\right\|_{\sigma}^2\leqslant \frac{B_{k+1}^2}{M}+8|\mA|\cdot\frac{\log(2/\delta)}{n}\cdot Q_{\max}^2\notag\\
&~~+ \sqrt{|\mA|}\cdot\left(16Q_{\max}(2B_{k+1}+2)\sqrt{\frac{2\log(2d)}{n}}+8Q_{\max}^2\sqrt{\frac{2\log(2c(B_{k+1}+1)^2/\delta)}{n}}\right),
\end{aligned}
\end{equation}
holds with probability at least $1-4|\mA|\delta$ over the sampling. In the next step, we are going to upper bound the $l_{\infty}$ norm of the function differences above. Notice that for a two-layer ReLU network $\widetilde{Q}_{k+1}$, its Lipschitz constant can be upper bounded by its path norm, which is because the ReLU activation function $\sigma(\cdot)$ itself is 1-Lipschitz continuous. Denote $L_{k+1}$ as the Lipschitz constant of $\widetilde{Q}_{k+1}-\Proj\left(T\widetilde{Q}_k, \mathcal C^{\oplus N}, \|\cdot\|_{\sigma}\right)$ on which we will analyze later. Notice that if $f$ is a continuous function with Lipschitz constant $L$, there is a relation on its $l_2$ norm and its $l_{\infty}$ norm. 
\begin{lemma}\label{lemma-lipschitz}
Suppose $f:\mathbb{R}^d\rightarrow \mathbb{R}$ is a continuous function with Lipschitz constant $L$, then we have:
\[\|f\|_2^2\geqslant \frac{\|f\|_{\infty}^{d+2}\cdot \pi^{d/2}}{3L^d\cdot d^2\Gamma\left(\frac{d}{2}+1\right)}.\]
\end{lemma}
\begin{proof}[Proof of Lemma \ref{lemma-lipschitz}]
By the definition of $\|f\|_{\infty}$, we know that for $\forall \varepsilon>0$, there exists $x_0\in\mathbb{R}^d$ such that $|f(x_0)|>\|f\|_{\infty}-\varepsilon$. Then for other $x\in\mathbb{R}^d$, it holds that
\[|f(x)|\geqslant \min\left(0, |f(x_0)|-L||x-x_0||\right),\]
since $|f(x)|\geqslant |f(x_0)|-|f(x_0)-f(x)|\geqslant |f(x_0)|-L\|x-x_0\|$. Therefore, the $l_2$ norm of $f$ can be lower bounded by:
\begin{equation*}
\begin{aligned}
\|f\|_2^2&\geqslant \int_{\mathbb{R}^d}\min\left(0, |f(x_0)|-L|x-x_0|\right)^2dx = \int_{B(x_0, |f(x_0)|/L)}(|f(x_0)|-L|x-x_0|)^2dx\\
&=\int_{0}^{|f(x_0)|/L}r^{d-1}\cdot(|f(x_0)|-Lr)^2 dr \cdot \frac{d\pi^{d/2}}{\Gamma\left(\frac{d}{2}+1\right)}\\
&= \frac{|f(x_0)|^{d+2}}{L^d}\cdot\frac{2\pi^{d/2}}{(d+1)(d+2)\Gamma\left(\frac{d}{2}+1\right)} \geqslant \frac{|f(x_0)|^{d+2}\cdot \pi^{d/2}}{3L^d\cdot d^2\Gamma\left(\frac{d}{2}+1\right)}.
\end{aligned}    
\end{equation*}
After making $\varepsilon\rightarrow 0$, we can conclude that:
\[\|f\|_2^2\geqslant \frac{\|f\|_{\infty}^{d+2}\cdot \pi^{d/2}}{3L^d\cdot d^2\Gamma\left(\frac{d}{2}+1\right)}.\]

\end{proof}
With the lemma above, we can finally get the upper bound for $\widetilde{Q}_{k+1}-\Proj\left(T\widetilde{Q}_k, \mathcal C^{\oplus N}, \|\cdot\|_{\sigma}\right)$. Assume that the density function of distribution $\sigma$ over $\mS\times\mA$ has a universal lower bound $c_{\sigma}$, then:
\begin{equation}
\label{eqn-lemma-non-3-9}
\begin{aligned}
\left\|\widetilde{Q}_{k+1}-\mathrm{Proj}\left(T\widetilde{Q}_k, \mathcal C^{\oplus N}, \|\cdot\|_{\sigma}\right)\right\|_{\sigma}^2\geqslant c_{\sigma}^2\cdot\left\|\widetilde{Q}_{k+1}-\mathrm{Proj}\left(T\widetilde{Q}_k, \mathcal C^{\oplus N},  \|\cdot\|_{\sigma}\right)\right\|_2^2,
\end{aligned}
\end{equation}
which leads to:
\begin{align}
\label{eqn-lemma-non-3-10}
&\left\|\widetilde{Q}_{k+1}-\Proj\left(T\widetilde{Q}_k, \mathcal C^{\oplus N}, \|\cdot\|_{\sigma}\right)\right\|_{\infty}^{d+2}\leqslant \frac{3L_{k+1}^d\cdot d^2 \Gamma\left(\frac{d}{2}+1\right)}{c_{\sigma}^2\cdot\pi^{d/2}}\cdot\Bigg[\frac{B_{k+1}^2}{M}+8|\mA|\cdot\frac{\log(2/\delta)}{n}\cdot Q_{\max}^2\notag\\
&~~+ \sqrt{|\mA|}\cdot\left(16Q_{\max}(2B_{k+1}+2)\sqrt{\frac{2\log(2d)}{n}}+8Q_{\max}^2\sqrt{\frac{2\log(2c(B_{k+1}+1)^2/\delta)}{n}}\right)\Bigg].
\end{align}
Notice that $\widetilde{Q}_{k+1}$ is a two-layer ReLU network with its path norm: $\max_{\ba}\|\widetilde{Q}_{k+1}(\cdot,\ba)\|_P\leqslant B_{k+1}$. For a two-layer ReLU network, its Lipschitz constant can be upper bounded by its path norm since its activation function $\sigma(\cdot)$ is 1-Lipschitz continuous, which leads to the fact that:
\[L_{k+1} = \mathrm{Lip}(\widetilde{Q}_{k+1})\leqslant \|\widetilde{Q}_{k+1}(\cdot,\ba)\|_P\leqslant B_{k+1}.\]
Next, we are going to determine the choice of the sequence of path norm upper bound $\{B_k\}$. Since $\widetilde{Q}_0\equiv 0$, we only need $B_0\geqslant 0$. From the proof above, we have already known that the sequence $\{B_k\}$ needs to satisfy: 
\[B_{k+1}\geqslant 4N\cdot\max_{i,a_i}\gamma\left(f_k^i(\cdot, a_i)\right),\]
where: $\Proj\left(T\widetilde{Q}_k, \mathcal{C}^{\oplus N}, \|\cdot\|_{\sigma}\right) = f_k^1(\gamma_1,a_1)+f_k^2(\gamma_2,a_2)+\ldots+f_k^N(\gamma_N,a_N)$. Notice that when $\widetilde{Q}_k\in \mathcal F(B_k)^{\oplus N}$, we have: for $\forall i\in[N], a_i\in\mathcal A$,
\[\|f_k^i(\cdot, a_i)\|_{\infty}\leqslant 2\|T\widetilde{Q}_k\|_{\infty}\leqslant 2(R_{\max}+\gamma\|\widetilde{Q}_k\|_{\infty})\leqslant 2(R_{\max}+\gamma\|\widetilde{Q}_k\|_P)
\leqslant 2(R_{\max}+N\gamma B_k).\]
Therefore, we only need to make $B_{k+1}=4N\cdot 2c(R_{\max}+N\gamma B_k)$ where $c$ is the upper bound of ratio between the spectral norm and the $l_{\infty}$ norm of a continuous function defined on $[0,1]^d$, which is a pure constant. So that we can guarantee that once $\widetilde{Q}_k\in \mathcal F(B_k)^{\oplus N}$, we have $B_{k+1}\geqslant 4N\cdot\max_{i,a_i}\gamma\left(f_k^i(\cdot, a_i)\right)$ holds. When the discount ratio $\gamma$ is small enough, we can make 
\[B_k = \frac{8NcR_{\max}}{1-4N^2\gamma}:= B>0,~~~\forall k\in [K].\] 
Now, the Equation (\ref{eqn-lemma-non-3-10}) becomes: with probability $p\geqslant 1-4|\mA|\delta$,
\begin{align}
\label{eqn-lemma-non-3-11}
&\left\|\widetilde{Q}_{k+1}-\Proj\left(T\widetilde{Q}_k, \mathcal C^{\oplus N}, \|\cdot\|_{\sigma}\right)\right\|_{\infty}^{d+2}\leqslant \frac{3B^d\cdot d^2 \Gamma\left(\frac{d}{2}+1\right)}{c_{\sigma}^2\cdot\pi^{d/2}}\cdot\Bigg[\frac{B^2}{M}+8|\mA|\cdot\frac{\log(2/\delta)}{n}\cdot Q_{\max}^2\notag\\
&~~+ \sqrt{|\mA|}\cdot\left(16Q_{\max}(2B+2)\sqrt{\frac{2\log(2d)}{n}}+8Q_{\max}^2\sqrt{\frac{2\log(2c(B+1)^2/\delta)}{n}}\right)\Bigg]
\end{align}
holds for $\forall k\in[K]$, which leads to the conclusion that:
\begin{align}
\label{eqn-lemma-non-3-12}
&\varepsilon_{\max}:= \max_{k\in[K]}\left\|\widetilde{Q}_{k+1}-\Proj\left(T\widetilde{Q}_k, \mathcal C^{\oplus N}, \|\cdot\|_{\sigma}\right)\right\|_{\infty} \leqslant c_1Bd\cdot\Bigg[\frac{B^2}{M}+8|\mA|\cdot\frac{\log(2/\delta)}{n}\cdot Q_{\max}^2\notag\\
&~~+ \sqrt{|\mA|}\cdot\left(16Q_{\max}(2B+2)\sqrt{\frac{2\log(2d)}{n}}+8Q_{\max}^2\sqrt{\frac{2\log(2c(B+1)^2/\delta)}{n}}\right)\Bigg]^{\frac{1}{d+2}}.
\end{align}
It comes to our conclusion after replacing $\delta$ with $\frac{\delta}{4|\mA|}$. 

\section{Existing Understanding on 2-layer ReLU Networks}
\label{appendix-e}
In this section, we introduce several important properties on neural networks, mainly on their approximation properties and generalization bounds. First, we study the two-layer ReLU networks. 

\subsection{Approximation Properties}
In this section, I will focus on the approximation of 2-layer ReLU networks to a target function. Assume $f^*:\Omega\rightarrow \mathbb{R}$ be the target function, where $\Omega = [-1,1]^{d}$, and $S=\{(\x_{i},y_{i})\}_{i=1}^{n}$ be the training set. Here the data points $\{\x_i\}_{i=1}^{n}$ are i.i.d samples drawn from an underlying distribution $\pi$ with $supp(\pi)\subset\Omega$, and $y_{i}=f^*(\x_{i})$. We aim to recover $f^*$ by fitting $S$ using a two-layer fully connected neural network with ReLU (rectified linear units) activation:
\[f(x;\theta)=\sum_{k=1}^{m}a_{k}\sigma(\tb_{k}\cdot\x+c_{k})\]
Here, function $\sigma(\cdot):\mathbb{R}\mapsto\mathbb{R}$ denotes the ReLU activation: $\sigma(t)=\max(0,t)$, $\tb_{k}\in R^{d}$ and the whole parameter set $\theta = \{(a_{k},\tb_{k},c_{k})\}_{k=1}^{m}$ is to be learned, and $m$ is the width of the network. In order to control the magnitude of learned network. We use the following scale-invariant norm.
\begin{Definition}
(Path norm \citep{neyshabur2015norm})For a two-layer ReLU network, the path norm is defined as:
\[\|\theta\|_{P} = \sum_{k=1}^{m}|a_{k}|(\|\tb_{k}\|_1+|c_{k}|)\]
\end{Definition}
\begin{Definition}
(Spectral norm) Given $f\in L^{2}(\Omega)$, denote by $F\in L^{2}(\mathbb{R}^{d})$ an extension of $f$ to $\mathbb{R}^{d}$. Let $\hat{F}$ be the Fourier transform of $F$, then:
\[f(\x) = \int_{\mathbb{R}^{d}}e^{i\langle\x,\w\rangle}\hat{F}(\omega)d\omega~~\forall\x\in\Omega\]
We define the spectral norm of $f$ by:
\[\gamma(f)=\inf\limits_{F\in L^{2}(\mathbb{R}^{d}),F|_{\Omega}=f|_{\Omega}}\int_{\mathbb{R}^d}\|\omega\|_1^2\cdot|\hat{F}(\omega)|d\omega\]
\end{Definition}
We also define $\hat{\gamma}(f)=\max\{\gamma(f),1\}$.

\begin{Assumption}
We consider target functions that are bounded and have finite spectral norm.
\[F_{s}=L^{2}(\Omega)\cap\{f(\x):\Omega\rightarrow\mathbb{R}|\gamma(f)<\infty, \|f\|_{\infty}\leqslant 1\}\]
We assume that $f^{*}\in F_s$.
\end{Assumption}
Since $\|f^{*}\|_{\infty}\leqslant 1$, we can truncate the network by $\tilde{f}(x)=\min\{|f(x)|,1\}\sign(f)$. By an abuse of notation, in the following we still use $f(x)$ to denote $\tilde{f}(x)$. Our goal is to minimize the generalization error (also known as population risk).
\[L(\theta)=\mathop{\mathbb{E}}\limits_{\x,y}[l(f(\x;\theta),y)]\]
However, practically, we only have to minimize the empirical risk
\[\hat{L}_{n}(\theta)=\frac{1}{n}\sum_{i=1}^{n}l(f(\x_i,\theta),y_{i})\]
Here, the generalization gap is defined as the difference between expected and empirical risk. The loss function is $l(y_{1},y_{2})=(y_{1}-y_{2})^2$ and that's why we analyze only regressive problems.

~\\
According to \citep{barron1993universal}, \citep{breiman1993hinging} and \citep{klusowski2016risk}, we can obtain the following approximation properties.
\begin{Lemma}
For any $F\in F_{s}$, one has the integral representation:
\[f(\x)-f(0)-\x\cdot\nabla f(0)=v\int_{\{-1,1\}\times[0,1]\times\mathbb{R}^{d}}h(\x;z,t,\w)dp(z,t,\w)\]
where:
\[p(z,t,\w)=|\hat{f}(\w)|\|\w\|_{1}^{2}|\cos(\|\w\|_{1}t-zb(\w))|/v\]
\[s(z,t,\w)=-\mathrm{sign}(\cos(\|\w\|_{1}t-zb(\w)))\]
\[h(\x,z,t,\w)=s(z,t,\w)(z\x\cdot\w/\|\w\|_{1}-t)_{+}\]
$v$ is the normalization constant such that $\int p(z,t,\w)dzdtd\w = 1$, which satisfies $v\leqslant 2\gamma(f)$.
\end{Lemma}
\begin{proof}
Since $f\in L^{2}(\mathbb{R}^{d})$, we have:
\[f(\x)-f(0)-\x\cdot\nabla f(0)=\int_{\mathbb{R}^{d}}(e^{i\w\cdot\x}-i\w\cdot\x-1)\hat{f}(\w)d\w\]
Note that the identity
\[-\int_{0}^{c}[(z-s)_{+}e^{is}+(-z-s)_{+}e^{-is}]ds=e^{iz}-iz-1\]
holds when $|z|\leqslant c$. Choosing $c=\|\w\|_{1}, z=\w\cdot\x$, we have;
\[|z|\leqslant \|\w\|_{1}\|\x\|_{\infty}\leqslant c\]
Let $s=\|\w\|_{1}t, 0\leqslant t\leqslant 1$, and $\hat{\w}=\w/\|\w\|_{1}$, we have:
\[-\|\w\|_{1}^{2}\int_{0}^{1}[(\hat{\w}\cdot\x-t)_{+}e^{i\|\w\|_{1}t}+(-\hat{\w}\cdot\x-t)_{+}e^{-i\|\w\|_{1}t}]dt=e^{i\w\cdot\x}-i\w\cdot\x-1.\]
Let $\hat{f}(\w)=e^{ib(\w)}|f(\w)|$, according to the two equations above:
\[f(\x)-f(0)-\x\cdot\nabla f(0)=\int_{\mathbb{R}^{d}}\int_{0}^{1}g(t,\w)dtd\w,\]
where:
\[g(t,\w) = -\|\w\|_{1}^{2}|\hat{f}(\w)|\cdot\left[(\hat{\w}\cdot\x-t)_{+}\cos(\|\w\|_{1}t+b(\w))+(-\hat{\w}\cdot\x-t)_{+}\cos(\|\w\|_{1}t-b(\w))\right].\]
Consider a density on $\{0,1\}\times[0,1]\times\mathbb{R}^{d}$ defined by:
\[p(z,t,\w)=|\hat{f}(\w)|\|\w\|_{1}^{2}|\cos(\|\w\|_{1}t-zb(\w))|/v\]
where the normalized constant $v$ is given by
\[v=\int_{\mathbb{R}^{d}}\int_{0}^{1}(|\cos(\|\w\|_{1}t+b(\w))|+|\cos(\|\w\|_{1}t-b(\w))|)dtd\w\]
Since $f\in F_{s}$, therefore:$v\leqslant 2\gamma(f)<+\infty$.
So, this density is well-defined. To simplify the notations, denote:
\[s(z,t,\w)=-\mathrm{sign}(\cos(\|\w\|_{1}t-zb(\w))),~~h(\x;z,t,\w)=s(z,t,\w)(z\hat{\w}\cdot\x-t)_{+}\]
Then we have
\[f(\x)-f(0)-\x\cdot\nabla f(0)=v\int_{\{-1,1\}\times[0,1]\times\mathbb{R}^{d}}h(\x;z,t,\w)dp(z,t,\w).\]
\end{proof}
For simplicity, in the following part, we assume $f(0)=0, \nabla f(0)=0$ because according to the equation above, we can use $f(\x)-f(0)-(\x\cdot\nabla f(0))_{+}+(-\x\cdot\nabla f(0))_{+}$ to replace $f(\x)$. This is a Monte-Carlo scheme. Therefore, we take $m$ samples $T_{m}=\{(z_{1},t_{1},\w_{1}),\cdots,(z_{m},t_{m},\w_{m})\}$ with $(z_{i},t_{i},\w_{i})$ randomly drawn from the probability density function $p(z,t,\w)$, and consider the empirical average $\hat{f}_{m}(\x)=\frac{v}{m}\sum_{k=1}^{m}h(\x;z_{i},t_{i},\w_{i})$, which is exactly a two-layer ReLU network of width $m$. The central limit theorem tells us that the approximation error:
\[\mathop{\mathbb{E}}\limits_{(z,t,\w)}[h(\x;z,t,\w)]-\frac{1}{m}\sum_{k=1}^{m}h(\x;z_{k},t_{k},\w_{k})\approx\sqrt{\frac{\mathrm{Var}_{(z,t,\w)}[h(\x;z,t,\w)]}{m}}\]
So what we have to do is bounding the variance on the right-hand side of the equation above.
\begin{theorem}\label{approx}
For any distribution $\pi$ with $supp(\pi)\subset\Omega$ and any $f\in F_{s}$, there exists a two-layer network $f(\x;\widetilde{\theta})$ of width $m$ such that:
\[\mathop{\mathbb{E}}\limits_{\x\sim\pi}|f(\x)-f(\x;\widetilde{\theta})|^{2}\leqslant \frac{16\gamma^{2}(f)}{m}\]
Furthermore, the path norm of the parameter $\widetilde{\theta}$ can be bounded by the spectral norm of the target function: $\|\theta\|_{P}\leqslant 4\gamma(f)$.
\end{theorem}
\begin{proof}
Let $\hat{f}_{m}(\x)=\frac{v}{m}\sum_{k=1}^{m}h(\x;z_{i},t_{i},\w_{i})$ be the Monte-Carlo estimator, then:
\begin{align}
\mathbb{E}_{T_{m}}\mathbb{E}_{\x}|f(\x)-\hat{f}_{m}(\x)|^{2} &= \mathbb{E}_{\x}\mathbb{E}_{T_{m}}|f(\x)-\hat{f}_{m}(\x)|^{2}\notag\\
&= \frac{v^{2}}{m}\mathbb{E}_{\x}(\mathbb{E}_{(z,t,\w)}[h^{2}(\x;z,t,\w)]-f^{2}(\x))\notag\\
&\leqslant \frac{v^{2}}{m}\mathbb{E}_{\x}\mathbb{E}_{(x,t,\w)}[h^{2}(\x;z,t,\w)]
\end{align}
For any fixed $\x$, the variance above can be bounded as:
\[\mathbb{E}_{(x,t,\w)}[h^{2}(\x;z,t,\w)]\leqslant\mathbb{E}_{(x,t,\w)}[(z\hat{\w}\cdot\x-t)_{+}^{2}]\leqslant\mathbb{E}_{(x,t,\w)}[(|\hat{\w}\cdot\x|+t)^{2}]\leqslant 4\]
Hence we have:
\[\mathbb{E}_{T_{m}}\mathbb{E}_{\x}|f(\x)-\hat{f}_{m}(\x)|^{2}\leqslant \frac{4v^{2}}{m}\leqslant\frac{16\gamma^{2}(f)}{m}\]
So we get the following conclusion: there exists a set of $T_{m}$, such that: $\mathbb{E}_{\x}|f-f_{m}|^{2}\leqslant\frac{16\gamma^{2}(f)}{m}$.
Notice the special structure of the Monte-Carlo estimator, we have: $|a_{k}|=\frac{v}{m},\|\tb_{k}\|_{1}=1,|c_{k}|\leqslant q$. Therefore, $\|\widetilde{\theta}\|_{P}\leqslant 2v\leqslant 4\gamma(f)$.
\end{proof}

\subsection{Generalization Properties}

\begin{Definition}
(Rademacher Complexity) Let $H$ be a hypothesis space. The Rademacher Complexity of $H$ with respect to samples $S=(z_{1},\cdots,z_{n})$ is defined as:
\[\hat{R}(H)=\frac{1}{n}\mathbb{E}_{\xi}\left[\sup_{h\in H}\sum_{i=1}^{n}h(z_{i})\xi_{i}\right]\]
where $\{\xi_{i}\}_{i=1}^{n}$ are independent random variables with probability $P(\xi_{i}=1)=P(\xi_{i}=-1)=\frac{1}{2}$
\end{Definition}
Before coming to the estimation of Rademacher Complexity, we need to introduce some fundamental properties.

\subsubsection{Basic Properties about Rademacher Complexity}
\begin{lemma}
\label{lemma-2}
For any $A\in \mathbb{R}^m$, scalar $c\in \mathbb{R}$, and vector $\ta_{0}\in \mathbb{R}^{m}$, we have:
\[R(\{c\ta+\ta_{0}:\ta\in A\})=|c|R(A)\]
\end{lemma}

Next, we are going to state several more important lemmas about Rademacher Complexity since they explain that the Rademacher Complexity of a finite set grows logarithmically with the size of the set.
\begin{lemma}[Massart Lemma] Let $A=\{\ta_{1},\ta_{2},\cdots,\ta_{N}\}$ be a finite set of vectors in $\mathbb{R}^{m}$. Then:
\[R(A)\leqslant \max\limits_{\ta\in A}\|\ta-\bar{\ta}\|\cdot\frac{\sqrt{2\log{N}}}{m}\]
Here: $\bar{\ta}=\frac{1}{N}\sum_{i=1}^{N}\ta_{i}$ is the average of all vectors in $A$.
\label{massart}
\end{lemma}
\begin{proof}
According to Lemma \ref{lemma-2}, we can assume $\bar{\ta}=0$ with loss of generality. Let $\lambda > 0$ and let $A'=\{\lambda \ta_{1},\lambda \ta_{2},\cdots,\lambda \ta_{N}\}$ where $\lambda$ is a positive scalar which remains to be determined. Then we calculate the upper bound of Rademacher Complexity of $A'$. 
\begin{align}
mR(A') &= \mathop{\mathbb{E}}\limits_{\sigma}\left[\max\limits_{\ta\in A'}<\sigma,\ta>\right] = \mathop{\mathbb{E}}\limits_{\sigma}\left[\log\left(\max\limits_{\ta\in A'}e^{<\sigma,\ta>}\right)\right]\notag\\
&\leqslant \mathop{\mathbb{E}}\limits_{\sigma}\left[\log\left(\sum_{\ta\in A'}e^{<\sigma,\ta>}\right)\right]\leqslant \log\left(\mathop{\mathbb{E}}\limits_{\sigma}\left[\sum_{\ta\in A'}e^{<\sigma,\ta>}\right]\right)\notag\\
&= \log\left(\sum_{\ta\in A'}\prod_{i=1}^{m}\mathop{\mathbb{E}}\limits_{\sigma_{i}}[e^{\sigma_{i}a_{i}}]\right)
\end{align}
Since:
\[\mathop{\mathbb{E}}\limits_{\sigma_{i}}[e^{\sigma_{i}a_{i}}] = \frac{1}{2}\left(\exp(a_{i})+\exp(-a_{i})\right)\leqslant \exp\left(\frac{a_{i}^{2}}{2}\right).\]
Therefore:
\begin{equation}
\begin{aligned}
mR(A')&\leqslant \log\left(\sum_{\ta\in A'}\prod_{i=1}^{m}\exp\left(\frac{a_{i}^{2}}{2}\right)\right) = \log\left(\sum_{\ta\in A'}\exp(\|\ta\|_{2}^{2}/2)\right)\\
&\leqslant \log(|A'|)+\max\limits_{\ta\in A'}(\|\ta\|_{2}^{2}/2)
\end{aligned}
\end{equation}
According to the definition of $A'$ and Lemma \ref{lemma-2}, we know that $R(A')=\lambda R(A)$. Then:
\[R(A)\leqslant \frac{\log(|A|)+\lambda^{2}\max\limits_{\ta\in A}(\|\ta\|_{2}^{2}/2)}{\lambda m}\]
Finally, set the optimal 
\[\lambda = \sqrt{\frac{2\log|A|}{\max\limits_{\ta\in A}(\|\ta\|_{2}^{2})}}\]
and we can come to our conclusion.
\end{proof}

The following shows that composing $A$ with a Lipschitz function will not blow up the Rademacher Complexity. And this is one of the most important conclusions about Rademacher Complexity.
\begin{lemma}[Contraction Lemma] 
For each $i\in[m]$, let $\phi_{i}:\mathbb{R}\rightarrow\mathbb{R}$ be a $\rho$-Lipschitz function, which means for all $x_{1},x_{2}\in\mathbb{R}$, we have:
\[|\phi_{i}(x_{1})-\phi_{i}(x_{2})|\leqslant\rho|x_{1}-x_{2}|\]
For $\ta\in\mathbb{R}^{m}$, let $\mathbf{\rho}(\ta)$ denote the vector $(\phi_{1}(a_{1}),\phi_{2}(a_{2}),\cdots,\phi_{m}(a_{m}))$ and $\mathbf{\phi}\circ A=\{\mathbf{\rho}(\ta):\ta\in A\}$. Then:
\[R(\mathbf{\phi}\circ A)\leqslant \rho R(A).\]
\label{contract}
\end{lemma}
\begin{proof}
For simplicity, we can assume $\rho = 1$. Otherwise, we can replace $\phi$ with $\phi'=\frac{1}{\rho}\phi$ and then use Lemma \ref{lemma-2} to prove our conclusion. Let:
\[A_{i}=\{(a_{1},\cdots,a_{i-1},\phi_{i}(a_{i}),a_{i+1},\cdots,a_{m}):\ta\in A\}\]
It is obvious that we only have to prove that for any set $A$ and all $i$, there holds:$R(A_{i})\leqslant R(A)$. Without loss of generality, we will prove that latter claim for $i = 1$ and to simplify notation, we omit the subscription of $\phi_{1}$. We have:
\begin{equation}
\label{eqn-contract}
\begin{aligned}
mR(A_{1})&= \mathop{\mathbb{E}}\limits_{\sigma}\left[\sup_{\ta\in A_{1}}\sum_{i=1}^{m}\sigma_{i}a_{i}\right]= \mathop{\mathbb{E}}\limits_{\sigma}\left[\sup_{\ta\in A}\sigma_{1}a_{1}+\sum_{i=2}^{m}\sigma_{i}a_{i}\right]\\
&=\frac{1}{2}\mathop{\mathbb{E}}\limits_{\sigma_{2},\cdots,\sigma_{m}}\left[\sup_{\ta\in A}\left(\phi(a_{1})+\sum_{i=2}^{m}\sigma_{i}a_{i}\right)+\sup_{\ta\in A}\left(-\phi(a_{1})+\sum_{i=2}^{m}\sigma_{i}a_{i}\right)\right]\\
&=\frac{1}{2}\mathop{\mathbb{E}}\limits_{\sigma_{2},\cdots,\sigma_{m}}\left[\sup_{\ta,\ta'\in A}\left(\phi(a_{1})-\phi(a_{1}')+\sum_{i=2}^{m}\sigma_{i}a_{i}+\sum_{i=2}^{m}\sigma_{i}a_{i}'\right)\right]\\
&\leqslant\frac{1}{2}\mathop{\mathbb{E}}\limits_{\sigma_{2},\cdots,\sigma_{m}}\left[\sup_{\ta,\ta'\in A}\left(|a_{1}-a_{1}'|+\sum_{i=2}^{m}\sigma_{i}a_{i}+\sum_{i=2}^{m}\sigma_{i}a_{i}'\right)\right]
\end{aligned}
\end{equation}
where in the last inequality, we used the Lipschitz condition of $\phi$. Next, we note that the absolute sign can be erased because both $\ta$ and $\ta'$ are from the same set $A$. Therefore, 
\[mR(A_{1})\leqslant\frac{1}{2}\mathop{\mathbb{E}}\limits_{\sigma_{2},\cdots,\sigma_{m}}\left[\sup_{\ta,\ta'\in A}\left(a_{1}-a_{1}'+\sum_{i=2}^{m}\sigma_{i}a_{i}+\sum_{i=2}^{m}\sigma_{i}a_{i}'\right)\right]\]
But using the same inequalities in Equation (\ref{eqn-contract}), it is easy to see that the right-hand side is equivalent to the occasion where $\phi_{1}=\mathrm{Id}$. Therefore, the right size exactly equals $mR(A)$, which comes to our conclusion.
\end{proof}

\begin{lemma}
\label{rc-L1}
Let $S=\{\x_{1},\x_{2},\cdots,\x_{m}\}$ be vectors in $\mathbb{R}^{n}$. Then, for the hypothesis class $H_1=\{x\mapsto \langle\w, x\rangle:~\|\w\|_2\leqslant 1\}$, we have:
\[R(H_{1}\circ S)\leqslant \max\limits_{i}\|\x_{i}\|_{\infty}\sqrt{\frac{2\log(2n)}{m}}\]
\end{lemma}
\begin{proof}
Using Holder's Inequality, we know that for any $\w,\tv$, we have: $\langle\w,\tv\rangle\leqslant \|\w\|_{1}\|\tv\|_{\infty}$. Therefore:
\begin{align}
mR(H_{1}\circ S) &= \mathop{\mathbb{E}}\limits_{\sigma}\left[\sup_{\ta\in H_{1}\circ S}\sum_{i=1}^{m}\sigma_{i}a_{i}\right]= \mathop{\mathbb{E}}\limits_{\sigma}\left[\sup_{\w:\|\w\|_{1}\leqslant 1}\sum_{i=1}^{m}\sigma_{i}\langle\w,\x_{i}\rangle\right]\notag\\
&= \mathop{\mathbb{E}}\limits_{\sigma}\left[\sup_{\w:\|\w\|_{1}\leqslant 1}\langle\w,\sum_{i=1}^{m}\sigma_{i}\x_{i}\rangle\right]\leqslant \mathop{\mathbb{E}}\limits_{\sigma}\left[\left\|\sum_{i=1}^{m}\sigma_{i}\x_{i}\right\|_{\infty}\right]
\end{align}
For $j\in [n]$, let $\tv_{j}=(x_{1,j},\cdots,x_{m,j})\in \mathbb{R}^{m}$. Note that: $\|\tv_{j}\|_{2}\leqslant \sqrt{m}\max_{i}\|\x_{i}\|_{\infty}$. Let $V=\{\tv_{1},\cdots,\tv_{n},-\tv_{1},\cdots,-\tv_{n}\}$. The right-hand side of Equation \ref{rc-L1} is $mR(V)$. Using Massart Lemma (Lemma \ref{massart}) we have that:
\[R(V)\leqslant \max\limits_{i}\|\x_{i}\|_{\infty}\sqrt{2\log(2n)/m}\]
\end{proof}
\begin{lemma}
\label{fund}
Assume that for all data points $s\in S$ and $h\in H$ where $H$ is a hypothesis set, we all have $l(h,z)\leqslant c$. Then: with probability of at least $1-\delta$, for $\forall h\in H$,
\[L_{D}(h)-L_{S}(h)\leqslant 2\underset{S'\sim D^{m}}{\mathbb{E}}R(l\circ H\circ S')+c\sqrt{\frac{2\ln{(2/\delta)}}{m}}\]
\end{lemma}
According to \citep{ma2018priori, don2020priori}, we can finally get an upper bound of Rademacher Complexity of 2-layer ReLU networks:
\begin{lemma}
\label{lemma-2layer}
Denote $F_{Q}=\{f_{m}(x;\theta):\mathbb{R}^{D}\rightarrow\mathbb{R}|\|\theta\|_{P}\leqslant Q\}$ be the set of two-layer ReLU networks with path norm bounded by $Q$, then we can bound its Rademacher Complexity.
\[R(F_{Q})\leqslant 2Q\sqrt{\frac{2\log(2D)}{n}}\]
\end{lemma}
\begin{proof}
To simplify the proof, we can assume $c_{k}=0$ without loss of generality. Otherwise, we can define $\tb_{k}=(\tb_{k}^{T},c_{k})^{T}, \x=(\x,1)^{T}$.
\begin{align}
n\hat{R}(F_{Q})&=\mathbb{E}_{\xi}\left[\sup_{\|\theta\|_{P}\leqslant Q}\sum_{i=1}^{n}\xi_{i}\sum_{k=1}^{m}a_{k}\|\tb_{k}\|_{1}\sigma(\hat{\tb}_{k}^{T}\x_{i})\right]\notag\\
&\leqslant\mathbb{E}_{\xi}\left[\sup_{\|\theta\|_{P}\leqslant Q,\|\tu_{k}\|_{1}=1}\sum_{i=1}^{n}\xi_{i}\sum_{k=1}^{m}a_{k}\|\tb_{k}\|_{1}\sigma(\tu_{k}^{T}\x_{i})\right]\notag\\
&\leqslant\mathbb{E}_{\xi}\left[\sup_{\|\theta\|_{P}\leqslant Q,\|\tu_{k}\|_{1}=1}\sum_{k=1}^{m}a_{k}\|\tb_{k}\|_{1}\sum_{i=1}^{n}\xi_{i}\sigma(\tu_{k}^{T}\x_{i})\right]\notag\\
&\leqslant\mathbb{E}_{\xi}\left[\sup_{\|\theta\|_{P}\leqslant Q}\sum_{k=1}^{m}|a_{k}\|\tb_{k}\|_{1}|\sup_{\|\tu\|_{1}=1}\left|\sum_{i=1}^{n}\xi_{i}\sigma(\tu^{T}\x_{i})\right|\right]\notag\\
&\leqslant Q\mathbb{E}_{\xi}\left[\sup_{\|\tu\|_{1}=1}
\left|\sum_{i=1}^{n}\xi_{i}\sigma(\tu^{T}\x_{i})\right|\right]\leqslant Q\mathbb{E}_{\xi}\left[\sup_{\|\tu\|_{1}\leqslant 1}\left|\sum_{i=1}^{n}\xi_{i}\sigma(\tu^{T}\x_{i})\right|\right].
\end{align}
Due to the symmetry, we have that:
\begin{align}
\mathbb{E}_{\xi}\left[\sup_{\|\tu\|_{1}\leqslant 1}|\sum_{i=1}^{n}\xi_{i}\sigma(\tu^{T}\x_{i})|\right] &\leqslant \mathbb{E}_{\xi}\left[\sup_{\|\tu\|_{1}\leqslant 1}\sum_{i=1}^{n}\xi_{i}\sigma(\tu^{T}\x_{i})+\sup_{\|\tu\|_{1}\leqslant 1}\sum_{i=1}^{n}-\xi_{i}\sigma(\tu^{T}\x_{i})\right]\notag\\
&= 2\mathbb{E}_{\xi}\left[\sup_{\|\tu\|_{1}\leqslant 1}\sum_{i=1}^{n}\xi_{i}\sigma(\tu^{T}\x_{i})\right]
\end{align}
Since the activation function $\sigma(\cdot)$ has Lipschitz constant 1. According to Lemma \ref{contract} and Lemma \ref{rc-L1}, we have:
\[R(F_{Q})\leqslant 2Q\sqrt{\frac{2\log(2D)}{n}}\]
which comes to our conclusion.
\end{proof}

Finally, we can combine Lemma \ref{lemma-2layer} with Lemma \ref{massart} and Lemma \ref{contract}, and obtain the following conclusion, which shows the generalization bound over the 2-layer ReLU networks. 

\begin{theorem}
\label{thm-generalization-2layer}
Suppose the loss function $l(\cdot, y)=(\cdot-y)^{2}$ is $\rho$-Lipschitz continuous and bounded by $B$. Then with probability at least $1-\delta$ over the choice of samples, we have:
\[\sup_{\|f\|_{P}\leqslant Q}|L(f)-\hat{L}_{n}(f)|\leqslant 4\rho Q\sqrt{\frac{2\log(2d)}{n}}+B\sqrt{\frac{2\log(2d/\delta)}{n}}\]
Here:
\[L(f)=\mathbb{E}_{(\x,y)\sim \pi}(f(\x)-y)^2,~\hat{L}_{n}(f)=\mathbb{E}_{\x\in S}(f(\x)-y)^2.\]
\end{theorem}

Then, by using the union bound, we conclude the following more general result.

\begin{theorem} [A posterior generalization bound] Assume the loss function $l(\cdot, y)$ is $\rho$-Lipschitz continuous and bounded by $B$. Then for any $\delta>0$, with probability at least $1-\delta$ over the choice the training set $S$, we have: for any two-layer ReLU network $f$, it holds that
\[|L(f)-\hat{L}_{n}(f)|\leqslant 4\rho (\|f\|_{P}+1)\sqrt{\frac{2\log(2d)}{n}}+B\sqrt{\frac{2\log(2c(\|f\|_{P}+1)^{2}/\delta)}{n}}\]
Here: $c=\sum_{k=1}^{+\infty}1/k^{2}=\pi^{2}/6$.
\label{thm-posterior}
\end{theorem}

\begin{proof}
Consider the decomposition of the full space $\mathcal F=\cup_{i=1}^{\infty}\mathcal F_{i}$, where $\mathcal F_{i}=\{f~\Big|\|f\|_{P}\leqslant i\}$. Let $\delta_{i}=\frac{\delta}{ci^2}$. According to Theorem \ref{thm-generalization-2layer}, if we fixed $i$ in advance, then with probability at least $1-\delta_i$ over the choice of $S$,
\[\sup_{\|f\|_P\leqslant i}|L(f)-\hat{L}_{n}(f)|\leqslant 4\rho i\sqrt{\frac{2\log(2d)}{n}}+B\sqrt{\frac{2\log(2d/\delta_{i})}{n}}\]
So the probability that there exists at least one $i$ to fail the inequality above is at most $\sum_{i=1}^{\infty}\delta_{i}=\delta$. In other words, with probability at least $1-\delta$, the inequality above holds for all $i$.
Given any two-layer ReLU network $f$ of width $M$, let $i_{0}=\lceil\|f\|_{P}\rceil$. Then:
\begin{align}
&|L(f)-\hat{L}_{n}(f)| \leqslant 4\rho i_{0}\sqrt{\frac{2\log(2d)}{n}} + B\sqrt{\frac{2\log(2c i_{0}^{2}/\delta)}{n}}\notag\\
&~~~~~~~~~~~\leqslant 4\rho (\|f\|_{P}+1)\sqrt{\frac{2\log(2d)}{n}} + B\sqrt{\frac{2\log(2c(\|f\|_{P}+1)^2/\delta)}{n}}
\end{align}
which comes to our conclusion.
\end{proof}

\end{document}